\documentclass{article}
\usepackage{microtype}
\usepackage{graphicx}
\usepackage{booktabs} 
\usepackage{hyperref}

\usepackage[accepted]{arxiv_version}

\usepackage[utf8]{inputenc}
\usepackage[english]{babel}
\usepackage{amsthm}
\usepackage{float}
\usepackage{amsmath}
\usepackage{listings}
\usepackage{marvosym}
\usepackage{color}
\usepackage{amssymb}
\usepackage{graphicx}
\usepackage{gensymb}
\usepackage{caption}
\usepackage{subcaption}
\usepackage{mathtools}
\usepackage[affil-it]{authblk}
\usepackage[mathscr]{euscript}
\usepackage{epigraph}
\usepackage{hyperref}
\usepackage{fancyhdr}
\usepackage{setspace}
\usepackage{dsfont}
\usepackage[ruled,vlined,algo2e]{algorithm2e}

\icmltitlerunning{Wasserstein GANs Work Because They Fail }

\DeclareMathOperator*{\argmax}{arg\,max}
\DeclareMathOperator*{\argmin}{arg\,min}

\theoremstyle{definition}
\newtheorem{definition}{Definition}[subsection]
\newtheorem*{definition*}{Definition}
\newtheorem{theorem}{Theorem}[subsection]
\newtheorem*{theorem*}{Theorem}

\newtheorem{lemma}[theorem]{Lemma}
\newtheorem*{lemma*}{Lemma}

\newtheorem{remark}{Remark}

\numberwithin{equation}{section}

\newcommand{\X}{\mathcal{X}}

\newcommand{\E}{\mathbb{E}}
\newcommand{\R}{\mathbb{R}}

\newcommand{\supp}{\text{supp}}
\newcommand{\norm}[1]{\left\lVert #1 \right\rVert}
\newcommand{\normLpz}[1]{\left\lVert #1 \right\rVert_{\text{L}}}

\newcommand{\loss}{\mathcal{L}}
\newcommand{\latent}{\mathcal{Z}}

\usepackage{xcolor}

\begin{document}

\twocolumn[
\icmltitle{Wasserstein GANs Work Because They Fail \\ (to Approximate the Wasserstein Distance)}

\icmlsetsymbol{equal}{*}

\begin{icmlauthorlist}
\icmlauthor{Jan Stanczuk}{cam}
\icmlauthor{Christian Etmann}{cam}
\icmlauthor{Lisa Maria Kreusser}{bath}
\icmlauthor{Carola-Bibiane Sch\"{o}nlieb}{cam}
\end{icmlauthorlist}

\icmlaffiliation{cam}{Cambridge Image Analysis Group, Department of Applied Mathematics and Theoretical Physics, University of Cambridge, Cambridge, United Kingdom}
\icmlaffiliation{bath}{Department of Mathematical Sciences, University of Bath, Bath, United Kingdom}

\icmlcorrespondingauthor{Jan Stanczuk}{js2164@cam.ac.uk}

\icmlkeywords{Machine Learning, ICML}

\vskip 0.3in
]

\printAffiliationsAndNotice{}  .

\begin{abstract}
Wasserstein GANs (WGANs) are based on the idea of minimising the Wasserstein distance between a real and a generated distribution. We provide an in-depth mathematical analysis of differences between the theoretical setup and the reality of training WGANs. In this work, we gather both theoretical and empirical evidence that the WGAN loss is not a meaningful approximation of the Wasserstein distance. In addition, we argue that the Wasserstein distance is not  a desirable loss function for deep generative models. We conclude that the success of WGANs can  be attributed to the failure to approximate the Wasserstein distance. 
\end{abstract}

\section{Introduction}
The Wasserstein GAN (WGAN), first introduced in \citep{WGAN}, is a framework for training Generative Adversarial Networks (GANs) by minimising the Wasserstein-1 distance (henceforth just called `the' Wasserstein distance) between a real and a generated distribution. The use of the Wasserstein distance was motivated as a remedy for shortcomings of the Jensen-Shannon (JS) divergence, which is implicitly used in vanilla GANs \citep{GAN}, but introduces some fundamental problems \citep{TrainingGANs}.

Over time,  practical implementations of WGANs have continuously been improved. Improvements include WGANs with gradient penalty (WGAN-GP) \citep{WGAN-GP} or, for specific cases only, so-called StyleGANs \citep{StyleGAN}. While the generative performance  of WGANs has gained wide-spread interest, other works \cite{MultiWasserstein, AR} train WGANs specifically to employ the trained discriminators (also called \emph{critics}) to estimate Wasserstein distances. 

In recent years, WGANs have become one of the most studied and successful deep generative models. Initially it was suggested that the success of WGANs can be attributed to the use of the Wasserstein distance. Many publications \cite{geomOT, DeepMultiW, WGAN-PCA, huang2019manifoldvalued, erdmann2018generating} (including some of our own prior work \cite{AR} and educational materials \cite{weng2019gan} still propagate or even rely on the assumption that WGANs are capable of approximating the Wasserstein distance accurately and that an accurate approximation is desirable.  However, the theoretical foundations of WGANs and the validity of the theoretical assumptions made in \citep{WGAN} have received little attention so far. In the studies by \citet{Pinetz,Mallasto}, the authors examine certain failures of the approximation of the correct Wasserstein distance via WGAN-GP.

In our work, we take a critical look at the original motivation of WGANs. We explore both theoretical and practical shortcomings of WGANs, and conclude that real-world implementations of WGANs should not be thought of as Wasserstein distance minimisers. We show that certain theoretical assumptions on WGANs are not satisfied in practise and infer that aiming for an accurate approximation of the Wasserstein distance with a WGAN loss is counterproductive and leads to worse results. We argue that the original success of WGAN-GP is likely to be the result of the regularisation of the discriminator (via Lipschitz constraints) and can be regarded as a carefully-chosen hyperparameter configuration rather than as a new loss function.

We  point out subtle differences between various possible notions of the ‘Wasserstein distance as a loss function’, including the `distributional' Wasserstein distance and the `batch' Wasserstein distance (for their definitions see Section \ref{sec:notation}).
The batch Wasserstein distance  is of interest in several works, see \cite{Mallasto, qpWGAN, learning_mb},  where it has been examined as a potential loss function for WGAN-based generative models.

Sample complexity estimates (compare Section \ref{sec:finite_sample_approx}) show  that estimating the true Wasserstein distance between distributions via batches requires prohibitive batch sizes. We demonstrate that even with a favourable sample complexity (obtained via entropic regularisation),  the generated samples of good batch Wasserstein estimators have a low fidelity.

For the case of Bernoulli measures, minimising  batch Wasserstein distances leads to noticeably different minimisers than minimising the actual Wasserstein distance between data-generating distributions \citep{Bellemare}. Moreover, there is a non-vanishing bias in sample estimates of the gradient of Wasserstein distances which can have an influence on gradient descent based learning algorithms. We take this analysis further and show empirically that false or undesirable minima occur not only in the case of Bernoulli measures, but also while learning common synthetic and benchmark distributions.

We show that a perfect generator – one that outputs actual samples from the data set – yields a significantly higher batch Wasserstein distance on average than a distribution concentrated on the centroids of geometric k-medians clustering. We prove that in a certain sense, these centroids comprise the batch that has the optimal Wasserstein distance to the data set. 

A variant of the batch Wasserstein distance has been suggested as a loss function for a generative model in \cite{learning_mb}, but experiments are only provided for 2D Gaussian data. Based on our theory we predict that when applied to image data a model based on the batch Wasserstein distance  results in blurry k-medians-like images which is also confirmed by our numerical experiments. We argue that the Wasserstein-1 distance is not a desirable loss function for deep generative models for image data due to the reliance on pixel-wise metrics, and we provide empirical evidence for this. We conclude that the success of WGANs can be attributed to the failure to approximate the Wasserstein distance.

While \citet{Mallasto} suggests using the $c$-transform as a more accurate batch Wasserstein estimator, a worse generative performance has been observed.
We demonstrate that the drop in performance is related to sample complexity and optimisation, but is also an issue of the Wasserstein distance itself due to the  euclidean distance as the underlying metric.

This analysis raises a natural question: If models like WGAN-GP (which generate high-fidelity samples) do not use a meaningful approximation of the Wasserstein distance,  why do they achieve said visual performance, often better than the vanilla GAN \citep{GAN}? The literature suggests two possible answers: Firstly, it has been proposed in \citep{DRAGAN} and \citep{Paths} that regularising the Lipschitz constraint of the discriminator may improve stability of GAN training \emph{regardless} of the statistical distance used as a loss function. Secondly, in \citep{GANsEqual}, a large-scale experiment has shown that vanilla GANs can achieve  similar performance to WGAN-GP if the right hyperparameters are chosen. Therefore, the original success of WGAN-GP might be due to a carefully-chosen hyperparameter configuration rather than because of the new loss function. Our observations agree with the points made in \citep{Paths} that GANs should not be understood as minimisers of some statistical distance. This suggests that the dynamics of the optimisation process based on alternating gradient updates need to be understood better and it is not sufficient to study the loss function or the optimal discriminator regime.

\subsection{Contributions}
The contributions of our work are as follows: 
\begin{enumerate}
    \item We provide a careful in-depth discussion on the modelling assumptions in \citep{WGAN}. We  specify the different ways in which the training of WGANs can (and in practice, does) fail to approximate Wasserstein distance estimators and we demonstrate these claims by providing  theoretical and empirical evidence.
    \item We point out subtle differences between various possible notions of the ‘Wasserstein distance as a loss function’. We show that for the batch Wasserstein distance, sample complexity issues fail to fully explain the failure (first noted by \citep{Mallasto}) of good batch Wasserstein estimators to generate high-fidelity samples.
    \item In addition to known results in \citep{Bellemare}, we demonstrate that the batch Wasserstein-1 distance is not even a desirable loss function for GANs. For this, we derive a new connection between the Wasserstein distance and clustering (geometric $k$-medians clustering), which results in  undesirable, low-fidelity samples, that nevertheless exhibit a low Wasserstein distance. We provide an experiment which shows that on the contrary, even a perfect generator (outputting actual data samples) yields a comparatively large Wasserstein distance.
    \item We argue that the fundamental problems of the Wasserstein-1 distance  stem from the underlying euclidean metric. We suggest that the failure to approximate the Wasserstein distance accurately  enables the generation of high-fidelity samples. In fact, the regularisation of the discriminator (in WGANs motivated by a Lipschitz constraint on the discriminator) helps generate good-looking samples, even when non-Wasserstein GANs are used.
\end{enumerate}

\subsection{Notation}\label{sec:notation}
In this work, we use the following notation. 
\begin{itemize}
    \item \textit{Empirical measures:}
    Given a probability distribution $p$, we denote the \textit{empirical measure} of $n$ samples by $p_n$, defined by $p_n : = \frac{1}{n}\sum_{i=1}^n \delta_{\substack{x_i}}$, where $x_i$ for $i=1,\ldots,n$ are independent and identically distributed (i.i.d.) samples from $p$.
Thus,  $p_n$ represents a mini-batch of $n$ samples from $p$ and we can consider $p_n$ as a distribution. %Notice that $p_m$ can be considered as fixed distribution if $x_i$ are fixed, or as a random variable if $x_i$ are random variables. 
%When regarding $p_m$ as a random variable we  denote its distribution by $P_m$. 
We write $x\sim p_n$ if $x$ is distributed according to $p_n$. The expectation of a given function $f$ satisfies $\mathbb{E}_{\substack{x \sim p_n}} [f(x)] =\int f(x) p_n(x) d x= \frac{1}{n} \sum_{i=1}^n f(x_i)$.
\item \textit{Set of empirical measures:} Given a probability distribution $p$, we denote the set of all empirical measures of $n$ samples drawn from $p$ by $\mathcal P_n$. We write $p_n\sim \mathcal P_n$ if we draw an empirical measure $p_n$ from the set $\mathcal P_n$.
\item \textit{Lipschitz continuity:}
    For a Lipschitz continuous function $f$, we denote its Lipschitz constant with respect to the euclidean norm (both in the domain and co-domain) by $\norm{f}_L$. 
    \item \textit{Wasserstein distance:} The Wasserstein distance is defined as
\begin{gather}
    W_1(p^\ast,p^\theta) := \inf_{\substack{\gamma\in \Gamma(p^\ast,p^\theta)}}  \mathbb{E}_{(x,y) \sim \gamma}[\norm{x - y}], \label{eq:primal_formulation}
\end{gather}
where the infimum is taken over all joint distributions $\gamma$ with marginals $p^\ast$ and $p^\theta$.
The above is referred to as the \textit{primal formulation} of Wasserstein distance. The Kantorovich-Rubenstein duality is given by
  \begin{align}
      W_1(p^\ast,p^\theta) = \max_{\substack{\norm{f}_L} \leq 1} \left( \mathbb{E}_{x \sim p^\ast}[f(x)] - \mathbb{E}_{x \sim p^\theta} [f(x)]\right). \label{eq:dual_formulation}
  \end{align}
  %where $\|f\|_L$ denotes the Lipschitz constant of $f$ with respect to the euclidean norm (both in the domain and co-domain). 
  This is referred to as the \textit{dual formulation}. The maximiser $f^\ast$ is called  \textit{Kantorovich's potential} between $p^\ast$ and $p^\theta$ and is determined (up to a constant) by $p^\ast$ and $p^\theta$. 
    \item \textit{Oracle estimator:}
    Let the empirical measures $p_n^\ast,p_n^\theta$ associated with the probability distributions $p^\ast,p^\theta$ be given. 
    We define the oracle estimator as
        \begin{gather*}
            {W_1^\ast}(p_n^\ast,p_n^\theta) =  \mathbb{E}_{x \sim p_n^\ast}[f^\ast(x)] - \mathbb{E}_{x \sim p_n^\theta} [f^\ast(x)],
        \end{gather*}
     where  
        \begin{align}\label{eq:kpotential}
        f^* \in \argmax_{\substack{\norm{f}_L \leq 1}} \left(\mathbb{E}_{x \sim p^\ast}[f(x)] - \mathbb{E}_{x \sim p^\theta} [f(x)] \right)
        \end{align}
        is  Kantorovich's potential.
    \item \textit{Mini-batch estimator:}
    Given empirical measures $p_n^\ast,p_n^\theta$, we define the mini-batch estimator as
        \begin{gather*}
        \hat{W}_1(p_n^\ast,p_n^\theta) =  \max_{\substack{\norm{f}_L}\leq 1}\left( \mathbb{E}_{x \sim p_n^\ast}[f(x)] - \mathbb{E}_{x \sim p_n^\theta} [f(x)]\right). 
        \end{gather*}
    \item \textit{Batch Wasserstein  and distributional Wasserstein distance}: Sometimes we wish to emphasise the difference between $W_1(p_n^\ast,p_n^\theta)$ and $W_1(p^\ast,p^\theta)$. In such cases we refer to the former as \textit{batch Wasserstein distance} and to the latter as \textit{distributional Wasserstein distance}.
\end{itemize}

\subsection{Outline}
This paper is structured as follows. In Section \ref{sec:background}, we give an overview on the original theory motivating the introduction of WGANs. In Section \ref{sec:estiamtion_of_w1_in_wgans}, we discuss how the Wasserstein distance is approximated in WGANs and distinguish two notions of Wasserstein distance as a loss function  in the literature: the \textit{batch Wasserstein distance} between mini-baches and the \textit{distributional Wasserstein distance}. We examine how well different WGANs approximate the loss function in Section \ref{sec:Approximation of the optimal discriminator}. We conclude that WGANs fail to approximate the distributional Wasserstein distance and that a better approximation of the Wasserstein distance between minibatches leads to worse generative performance. In Section~\ref{sec:finite_sample_approx}, we explore how sample complexity makes the efficient approximation of distributional Wasserstein distance impossible. We further investigate false minima of the batch Wasserstein distance and their connection to clustering. In section \ref{sec:fundamental_problems}, we discuss  fundamental issues of the Wasserstein distance as a loss function for image data stemming from the fact that it is based on the pixelwise $L_2$ distance. In section \ref{sec: WGAN_space_of_GANs}, we discuss possible explanations for the initially reported success of WGAN in light of the failure to approximate Wasserstein distance.

\section{Original Motivations for Wasserstein GAN}
\label{sec:background}
\subsection{Theoretical formulation of GANs}
Generative adversarial networks (GANs) were introduced in \citep{GAN} as a new framework for generative models.
A GAN consists of two neural networks: the generator $G_\theta: \latent \rightarrow \X$ and the discriminator  $D_\alpha: \X \rightarrow \R$ which compete against each other. Here, $\latent$ denotes the latent space, $\X$ is the data space and  $\theta,\alpha$ denote the parameters of the respective networks. The space $\latent$ is usually endowed with a multivariate Gaussian distribution $p_z$. For $z\in \latent$ with $z\sim p_z$, the outputs of the generator $G_\theta(z)$ form a distribution which we call the \emph{generator distribution} and denote by $p
^\theta$. The generator learns to produce samples which resemble the data from a \emph{target distribution} $p^*$, while the discriminator is trying to distinguish fake from real data (by assigning an estimated probability that $D_\alpha(x)$ is `real'  rather than generated data). In the context of WGANs, the discriminator is often called `critic'. Hence, we train $D_\alpha$ to maximize the probability of assigning the correct label to both training examples and samples from $G_\theta$, while we train $G_\theta$ to minimise the discrepancy between the generated samples and  data. Formally, given a value function $V(G_\theta, D_\alpha)$ the optimisation objective is of the form
\begin{gather*}
  \min_{\theta} \max_{\alpha} V(G_\theta, D_\alpha).
\end{gather*}

\subsection{Optimal discriminator dynamics}
A common approach to analysing the training of GANs is the so-called \textit{optimal discriminator dynamics}. In the optimal discriminator dynamics approach, we define $F(G_\theta) := \max_\alpha V(G_\theta,D_\alpha)$ and analyse GANs as a minimisation problem (rather than a mini-max problem):
\begin{gather*}
  \min_\theta \max_\alpha V(G_\theta,D_\alpha) = \min_\theta F(G_\theta).
\end{gather*}
This approach to modelling GAN dynamics relies on what we call the \textit{optimal discriminator assumption} (ODA), i.e.\ that after each update of the generator, we assume that the best possible discriminator was picked. This is not the case in practice, as we will discuss in later sections. 

If we choose the right value function,  we can interpret GAN training as a minimisation of a statistical divergence between a target and a generated distribution  under the ODA. For example in \citep{GAN}, it has been shown that vanilla GAN's value function 
\begin{align*}
    &V(G_\theta,D_\alpha) \\  
    = &\E_{x \sim p^*}[\log D_\alpha(x)] + \E_{\substack{z \sim p_z}}[\log (1-D_\alpha(G_\theta(z)))]
\end{align*}
induces the minimisation of the Jensen-Shannon divergence 
$$ JS(p^\ast, p^\theta) = \frac{1}{2} \bigg(KL\big(p^\ast \big|\big| \frac{p^\ast + p^\theta}{2} \big) + KL\big(p^\theta \big|\big| \frac{p^\ast + p^\theta}{2}\big)\bigg) $$ between the real distribution $p^\ast$ and the generator distribution $p^\theta$ where $KL$ denotes the Kullback-Leibler  divergence.
\subsection{Choosing the right divergence}
A rigorous mathematical analysis of the vanilla GAN's optimal discriminator dynamics has been performed in \citep{TrainingGANs}. The authors prove that for the vanilla GAN's value function, an accurate approximation of the optimal discriminator  leads to vanishing gradients passed to the generator (i.e.\ $\nabla_\theta V(G_\theta, D_\alpha) \rightarrow 0$ as $D_\alpha$ approaches the optimal discriminator $D^\ast$). This problem can be traced back to the fact that the JS divergence is maximised whenever  two distributions have disjoint supports. To address this issue, the authors in \cite{WGAN} suggest to replace the JS divergence by the Wasserstein distance $W_1$ which decreases smoothly as the supports of the distributions converge to each other.
In order to apply the Wasserstein distance to GAN training, the authors use the Kantorovich-Rubenstein duality \eqref{eq:dual_formulation}
  and redefine the GAN objective function as 
  \begin{align}\label{eq:loss}
  V(G_\theta,D_\alpha) = \mathbb{E}_{x \sim p^*}[D_\alpha(x)] - \mathbb{E}_{\substack{z \sim p_z}} [D_\alpha(G_\theta(z))].
  \end{align}
  The mini-max objective can be rewritten as
\begin{gather} \label{eq:minimaxW}
\min_\theta \max_{\substack{\norm{D_\alpha}_L \leq 1}} V(G_\theta,D_\alpha) = \min_{\theta} W_1(p^\ast, p^\theta),
\end{gather}
which shows that the 1-Lipschitz functions act as the class of possible discriminators. 

\section{Estimation of $W_1$ in WGANs}
\label{sec:estiamtion_of_w1_in_wgans}
\subsection{WGAN-GP Algorithm}
The exact computation of 
\begin{align}
\label{KR-Duality}
 W_1(p^\ast, p^\theta) = \max_{\substack{\norm{f}_L} \leq 1} \left(\mathbb{E}_{x \sim p^\ast}[f(x)] - \mathbb{E}_{x \sim p^\theta} [f(x)] \right)
\end{align}
is in practice impossible for two reasons. Firstly, it is computationally impossible to optimise over the set of all 1-Lipschitz functions accurately. Secondly, we do not have access to the full measures $p^\ast$ and $p^\theta$, but  only to finite samples from each of them. Therefore, the Wasserstein distance has to be approximated via some tractable loss function in WGANs.
There are many suggestions for loss functions in the literature (e.g.\ \citep{WGAN,WGAN-GP,SNGAN}). We  focus on the most prominent approximation scheme, introduced in \cite{WGAN-GP}. The function $f$ in the duality formula (\ref{KR-Duality}) is replaced by a neural network $D_\alpha$, which is then trained to maximise (\ref{KR-Duality}). Moreover the network is (at least approximately) constrained to be a 1-Lipschitz function. For this reason a regularisation term called \textit{gradient penalty} is incorporated in the loss function. More precisely, let $p^\ast_n$ and $p^\theta_n$ denote the empirical measures associated with measures $p^\ast$ and $p^\theta$ for the $n$ samples $(x_i)_{i=1}^n$ and $(\tilde{x}_i)_{i=1}^n$, respectively. Define
\begin{align}
    \mathcal V(D_\alpha, p_n^*, p_n^\theta) :&= \mathbb{E}_{x \sim p^\ast_n}[D_\alpha(x)] - \mathbb{E}_{x \sim p^\theta_n} [D_\alpha(x)],\label{eq:lossapprox} \\
    \mathcal{R}(D_\alpha, p^\ast_n, p^\theta_n) :&= \mathbb{E}_{x \sim \tau}[
    (\norm{\nabla_x D_\alpha(x)} - 1)^2],\label{eq:res}
\end{align}
where $\tau := \tau(p^\ast_n, p^\theta_n)$ is defined as %follows: let $(x_i)_{i=1}^m$ and $(\tilde{x}_i)_{i=1}^m$ be points in $p^\ast_m$ and $p^\theta_m$ respectively. $\tau$ is a
the uniform distribution on the lines connecting $x_i$ with $\tilde{x}_i$ for $i = 1, ... , m$. Note that \eqref{eq:lossapprox} can be regarded as an approximation of \eqref{eq:loss}, while \eqref{eq:res} enforces the gradient penalty.

Then one can optimise an approximation of the mini-max objective in \eqref{eq:minimaxW} in an iterative fashion.  First sample batches $ p_n^*, p_n^\theta$ and make a gradient ascent step with respect to the discriminator loss function $\loss_D(\alpha) := \mathcal V(D_\alpha, p_n^*, p_n^\theta) - \lambda \mathcal{R}(D_\alpha, p^\ast_n, p^\theta_n)$. Repeat this process $N_D$ times (to approximate $D_{\substack{\alpha^\ast}} = \arg\max_{\substack{\norm{D_\alpha}_L \leq 1}} V(G_\theta,D_\alpha)$). Then sample new batches $ p_n^*, p_n^\theta$ and make a gradient descent step with respect to the generator loss  $\loss_G(\theta) := \mathcal V(D_\alpha, p^*_n, p^\theta_n)$. Repeat the whole procedure $N_G$ times. The WGAN-GP is described in pseudo-code in the Algorithm \ref{alg:WGAN-GP}.

\begin{remark}
In \eqref{eq:lossapprox} we use a different notation for the value function $\mathcal V$ instead of $V$ in \eqref{eq:loss}. First notice that $V(G_\theta, D_\alpha)$ in \eqref{eq:loss} depends on $G_\theta$ only through $p^\theta$ and hence we may regard $V$ as a function of $D_\alpha$ and $p^\theta$. Sometimes we want to refer explicitly to the value function evaluated using randomly sampled mini-batches $p^\ast_n$ and $p^\theta_n$. In this case we shall write $\mathcal V(D_\alpha, p^\ast_n, p^\theta_n)$ as in \eqref{eq:lossapprox}.
\end{remark}

\begin{algorithm2e}
   \SetAlgoLined
    \DontPrintSemicolon
    \KwIn{$N_G$ -  number of generator updates, $N_D$ - number of discriminator updates per one generator update, $\lambda$ - gradient penalty regularisation parameter}
     \For{$N_G$ iterations}{
        \For{$N_D$ iterations}{
        Sample a batch $p^*_n$ from $p^\ast$ \;
        Sample a batch $p^\theta_n$ from $p^\theta$ \;
        Ascent $\alpha$\ wrt. $ \loss_D(\alpha) := \mathcal V(D_\alpha, p^*_n, p^\theta_n) - \lambda \mathcal{R}(D_\alpha, p^\ast_n, p^\theta_n)$ \;
        }
    Sample a batch $p^*_n$ from $p^\ast$ \;
    Sample a batch $p^\theta_n$ from $p^\theta$ \;
    Descent $\theta$ wrt. $\loss_G(\theta) := \mathcal V(D_\alpha, p^*_n, p^\theta_n)$
    }
\caption{WGAN-GP}
\label{alg:WGAN-GP}
\end{algorithm2e}

\begin{remark} 
    In Algorithm \ref{alg:WGAN-GP} we could have removed the second sampling from $p^*_n$ and descent wrt. $-\mathbb{E}_{x \sim p^\theta_n} [D_\alpha(x)]$ instead of $\mathcal V(D_\alpha, p^*_n, p^\theta_n)$. This would result in the same minimiser $\theta^\ast$, but in such case $\loss_G(\theta)$ would not approximate $W_1$.
\end{remark}

\subsection{c-transform WGAN}
 An approximation scheme based on $c$-transform has been proposed in \cite{Mallasto} which gives a  more accurate approximation of $W_1$ between mini-batches than WGAN-GP.  The main idea of their approach is to replace the Kantorovich-Rubenstein duality with the following so-called \textit{weak duality} formula:

\begin{theorem}[Weak Duality,  \cite{Mallasto}]\label{th:weakduality}
For probability distributions $p^\ast,p^\theta$, we have
    \begin{gather*}
        W_1(p^\ast,p^\theta) = \sup_{\substack{f \in \mathcal{C}_b}} \left(\mathbb{E}_{\substack{x \sim p^\ast}}[f(x)] + \mathbb{E}_{\substack{x \sim p^\theta}}[f^c(x)]\right),
    \end{gather*}
    where the supremum is taken over the space $\mathcal{C}_b$ of all continuous bounded functions such that $f\in \mathcal{C}_b$ satisfies $f: \mathcal{X} \rightarrow \mathbb{R}$ and $f^c(x) : = \sup_y \{f(y) - \norm{x - y} \}$ is its \textit{c-transform} of $f$.
\end{theorem}
Note that the $c$-transform of a 1-Lipschitz function $f$ is given by $f^c = -f$. Hence, $f^c$ is easy to compute for 1-Lipschitz functions. However, note that the optimisation is over the space of $\mathcal C_b$. 

In the $c$-transform WGAN, the authors use the weak duality (Theorem \ref{th:weakduality}) instead of Kantorovich-Rubinstein duality. This allows for the optimisation of the discriminator to be unconstrained, but introduces an approximate $c$-transform in the objective which reads
 \begin{align*}%\label{eq:loss}
        V(G_\theta,D_\alpha) = \mathbb{E}_{\substack{x \sim p^\ast}}[f(x)] + \mathbb{E}_{\substack{x \sim p^\theta}}[f^c(x)]
  \end{align*}
  The mini-max objective can be rewritten as
\begin{gather*} %\label{eq:minimaxW}
\min_\theta \max_{\alpha} V(G_\theta,D_\alpha) = \min_{\theta} W_1(p^\ast, p^\theta).
\end{gather*}
Similarly to the WGAN-GP in Section \ref{sec:estiamtion_of_w1_in_wgans}, we consider the loss approximation
\begin{align}\label{eq:lossapproxctransf}
    \mathcal V(D_\alpha, p_n^*, p_n^\theta) :&= \mathbb{E}_{x \sim p^\ast_n}[D_\alpha(x)] + \mathbb{E}_{x \sim p^\theta_n} [\hat{D}^c_\alpha(x)],
\end{align}
where $\hat{D}^c_\alpha$ is an approximation to $c$-transform of $D_\alpha$ given as $\hat{D}^c_\alpha(x) := \min_{y \in \text{supp}(p^\theta_n)} \norm{x-y} - D_\alpha(y)$.

The algorithm for the $c$-transform WGAN is as in Algorithm \ref{alg:WGAN-GP}, but with $\loss_D(\alpha) = \loss_G(\theta) = \mathcal V(D_\alpha, p_n^*, p_n^\theta)$ defined in \eqref{eq:lossapproxctransf}

\subsection{The oracle estimator}

The main idea behind the WGAN-GP algorithm is that $D_\alpha$, optimised in the inner loop, approximates  Kantorovich's potential between $p^\ast$ and $p^\theta$ in \eqref{eq:kpotential}. As a result the loss function of the generator  approximates the  \textit{oracle estimator} of the Wasserstein distance
\begin{gather*}
    {W_1^\ast}(p_n^\ast,p_n^\theta) =  \mathbb{E}_{x \sim p_n^\ast}[f^\ast(x)] - \mathbb{E}_{x \sim p_n^\theta} [f^\ast(x)],     
\end{gather*}
 where $$f^* \in \argmax_{\substack{\norm{f}_L \leq 1}} \left(\mathbb{E}_{x \sim p^\ast}[f(x)] - \mathbb{E}_{x \sim p^\theta} [f(x)]\right).$$ 

From the above discussion, we can conclude that there are two sources of error in the  approximation of the Wasserstein distance:
\begin{enumerate}
    \item Not learning the optimal discriminator exactly.
    \item Estimation of the expectations based on finite samples.
\end{enumerate}

We  discuss the impact of each source of error in the following sections. Moreover, we notice that even if we approximate the Wasserstein distance perfectly we still need to perform a non-convex optimisation via a stochastic gradient descent based learning algorithm on $W_1(p^\ast, p^\theta)$ in order to successfully train a GAN.

\subsection{The batch estimator}
Recently, some researchers have examined how well the loss function of WGAN approximates the distance between random mini-batches \cite{Mallasto}. More precisely, instead of approximating the oracle estimator they suggest that the loss function of WGAN should approximate the  \textit{batch estimator}
\begin{gather*}
    \hat{W}_1(p_n^\ast,p_n^\theta) =  \max_{\substack{\norm{f}_L}\leq 1}\left( \mathbb{E}_{x \sim p_n^\ast}[f(x)] - \mathbb{E}_{x \sim p_n^\theta} [f(x)] \right).
\end{gather*}
Here we have following sources of error:
\begin{enumerate}
    \item Not learning the optimal discriminator exactly.
    \item Fitting the discriminator to $p_n^\ast$ and $p_n^\theta$ instead of $p^\ast$ and $p^\theta$ (sample complexity).
\end{enumerate}

Using Theorem \ref{th:weakduality}, the batch estimator can be written as
\begin{align*}
     \hat{W}_1(p_n^\ast,p_n^\theta) = \sup_{\substack{f \in \mathcal{C}_b}} \left(\mathbb{E}_{\substack{x \sim p_n^\ast}}[f(x)] + \mathbb{E}_{\substack{x \sim p_n^\theta}}[f^c(x)]\right).
\end{align*}

\section{Approximation of the optimal discriminator}
\label{sec:Approximation of the optimal discriminator}

In the following, we discuss how accurately the optimal discriminator $D_\alpha$ is approximated in the different methods for the estimation of the Wasserstein distance. In other words, we investigate whether the loss function of WGAN-GP $\loss_G$ satisfies the approximations $\loss_G(\theta) \approx W_1^\ast(p^\ast, p^\theta)$ and $\loss_G(\theta) \approx \hat{W}_1(p^\ast, p^\theta)$. This question has been explored in two recent works by \citet{Mallasto} and \citet{Pinetz}, but our experiments differ significantly. The subtle, but crucial differences are explained in detail in the Appendix \ref{sec:prevexp}.

\begin{figure}[htb]
    \centering
    \includegraphics[scale=.25]{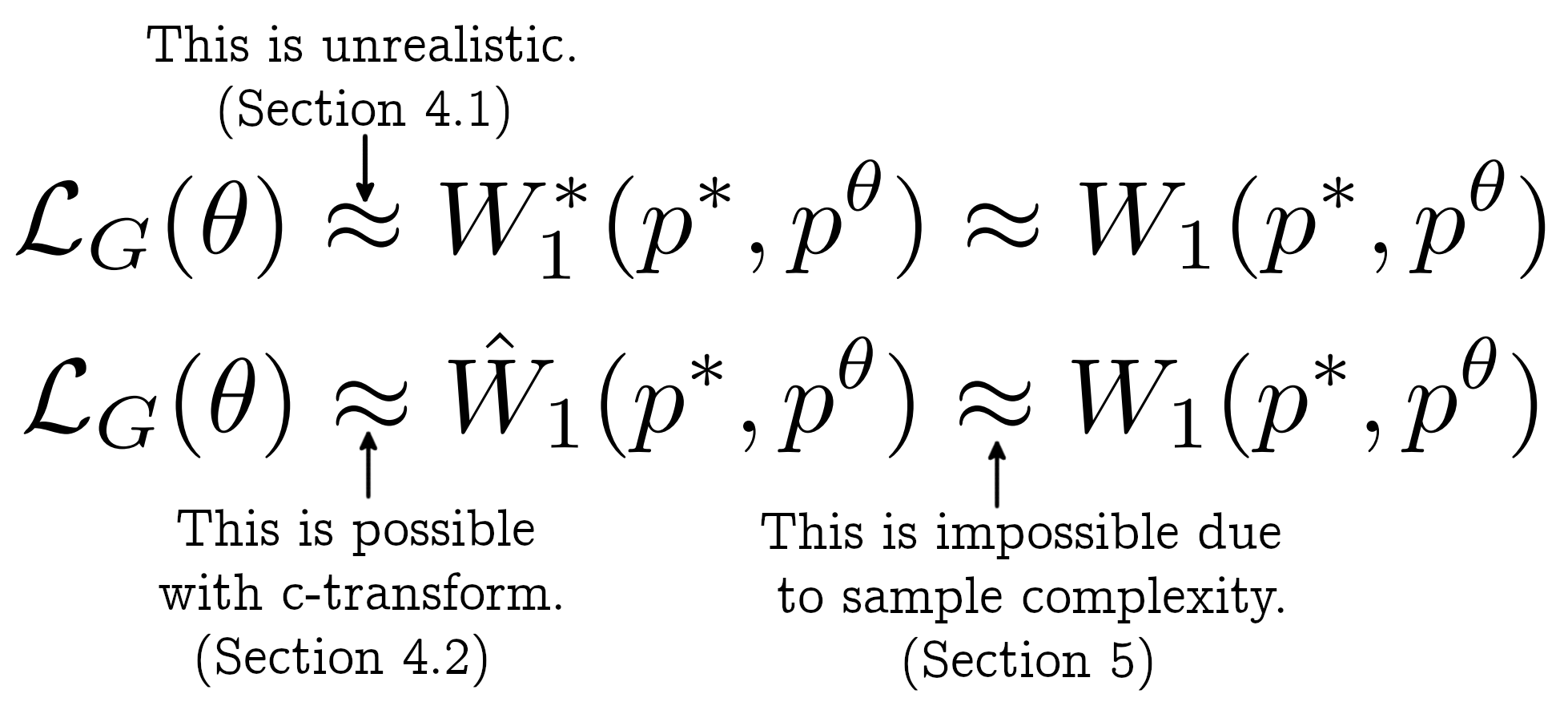}
    \caption{Overview of the desired approximations of $\loss_G$.}
    \label{fig:wganapprox}
\end{figure}

We show the following relations which are summarised in Figure \ref{fig:wganapprox}:
\begin{itemize}
    \item The loss function $\loss_G$ %of WGAN-GP 
    fails to approximate the oracle estimator ${W_1^\ast}(p^\ast, p^\theta)$ because the inner loop of Algorithm \ref{alg:WGAN-GP} fails to capture the optimal discriminator (Section \ref{sec:Does WGAN-GP provide a good oracle ?}).
    \item The loss function $\loss_G$
    %of WGAN 
    can approximate the batch estimator $\hat{W}_1(p^\ast, p^\theta)$ when trained using the $c$-transform, but $\hat{W}_1(p^\ast, p^\theta)$ is not a good approximation of the distribution level Wasserstein distance $W_1(p^\ast, p^\theta)$ (Section \ref{sec:approximation_of_batch_estimator}).
    \item The batch estimator $\hat{W}_1(p^\ast, p^\theta)$ of the Wasserstein distance $W_1(p^\ast, p^\theta)$ is not a desirable loss function for a generative model (Section \ref{sec:finite_sample_approx}).
    \item The close connection of the Wasserstein distance to the pixelwise $L_2$ norm causes fundamental issues when applying the Wasserstein distance to image data (Section~\ref{sec:fundamental_problems}).
\end{itemize}

%\CE{Is this enumeration not a repetition of the 'contributions' section from the introduction? Not sure whether we should tailor it more to this section of make it more different somehow. }\LK{this comment was originally at the end of section 3}

\subsection{Approximation of the oracle estimator}\label{sec:approximation_of_oracle}
\label{sec:Does WGAN-GP provide a good oracle ?}
First, we examine whether
\begin{gather}\label{eq:oracleapprox}
    \loss_G(\theta) \approx W_1^\ast(p^\ast, p^\theta)
\end{gather} 
is a valid approximation. This is the case if and only if the inner loop of the WGAN-GP algorithm \ref{alg:WGAN-GP}, also called the \textit{discriminator loop}, computes a good approximation $D_\alpha$ of  Kantorovich's potential $f^\ast$ in \eqref{eq:kpotential}.

\subsubsection{Approximation for fixed, finitely supported distributions}\label{sec:approxfixed}

 To examine whether \eqref{eq:oracleapprox} is satisfied in practice, we design the following experiment, summarised in Algorithm~\ref{alg:oracleapprox}. 
We pick two large finitely supported distributions $p^\ast$ and $p^\theta$, each consisting of $10$K images  from CIFAR-10 \cite{CIFAR}.  Then we sample mini-batches (of size $n=64$) from $p^\ast,p^\theta$ and maximise $\loss_D(\alpha)$. This is exactly the same procedure as in the WGAN-GP training in Algorithm \ref{alg:WGAN-GP} except that  both measures are static (as if the generator in Algorithm \ref{alg:WGAN-GP} was frozen). We consider $N=300K$ updates for $D_\alpha$. At the end, we check if the approximation 
\begin{align*}
W_1^D(p^\ast,p^\theta)&=\mathcal V(D_\alpha, p^\ast, p^\theta)\\&=\mathbb{E}_{x \sim p^\ast}[D_\alpha(x)] - \mathbb{E}_{x \sim p^\theta}[D_\alpha(x)]
\end{align*}
is close to $W_1(p^\ast,p^\theta)$. Note that we return  $\mathcal V(D_\alpha, p^\ast, p^\theta)$ and not $\mathcal V(D_\alpha, p^\ast, p^\theta)-\lambda \mathcal R(D_\alpha, p^\ast, p^\theta)$ for $W_1^D(p^\ast,p^\theta)$ in case the Lipschitz penality is not well satisfied.

Since the distributions are finitely supported, $W_1(p^\ast,p^\theta)$ can be obtained by solving a linear program (LP) as in \cite{POT}. Note that superficially similar experiment to Algorithm \ref{alg:oracleapprox} have been performed in \citep{Pinetz} and \cite{Mallasto}. Subtle, but crucial differences in the design of the experiment are discussed in  Appendix~\ref{sec:prevexp}.

\begin{algorithm2e}
    \SetAlgoLined
    \DontPrintSemicolon
    \For{$N$ iterations}{
        Sample a batch $p_n^\ast$ from $p^\ast$ \;
        Sample a batch $p_n^\theta$ from $p^\theta$ \;
        Ascent step on $D_\alpha$ wrt. $\mathcal V(D_\alpha, p_n^\ast, p_n^\theta) - \lambda \mathcal{R}(D_\alpha, p^\ast_n, p^\theta_n)$\;
    }
    $W_1^D(p^\ast,p^\theta) \gets \mathbb{E}_{x \sim p^\ast}[D_\alpha(x)] - \mathbb{E}_{x \sim p^\theta}[D_\alpha(x)]$ \;
    $W_1(p^\ast,p^\theta) \gets $ Solution of LP for $p^\ast,p^\theta$\;
    Compare $W_1^D(p^\ast,p^\theta)$ and $W_1(p^\ast,p^\theta)$
\caption{Quality of oracle estimation for static distributions.\label{alg:oracleapprox}
}
\end{algorithm2e}

Notice that if $D_\alpha \approx f^\ast$ then $W_1^D(p^\ast,p^\theta) \approx W_1(p^\ast,p^\theta)$.  Therefore we can assess the quality of the approximation of the optimal discriminator $D_\alpha$ in WGAN-GP by examining how much $W_1^D(p^\ast,p^\theta)$ differs from $W_1(p^\ast,p^\theta)$.

\begin{remark}\label{rem:lipschitznormalisation}
    As pointed out in  \cite{WGAN} if we optimise over the set of $K$-Lipschitz functions instead of the set of $1$-Lipschitz functions, then the estimate of Wasserstein distance  scales by the same factor, i.e.
    \begin{gather*}
        \max_{\substack{\normLpz{f}\leq K}}\left( \mathbb{E}_{x \sim p^\ast}[f(x)] - \mathbb{E}_{x \sim p^\theta}[f(x)]\right) = K W_1(p^\ast,p^\theta).
    \end{gather*}
    Since the 1-Lipschitz continuity of $D_\alpha$ is  only approximately enforced in WGAN-GP, a normalisation by the Lipschitz constant of $D_\alpha$ should be considered. We define the lower bound of $\normLpz{D_\alpha}$ by $\hat{L}(D_\alpha) := \max_{x \in \text{supp}(\tau)} \norm{\nabla D_\alpha(x)}$ and consider the \textit{normalised Wasserstein estimate} as $W^D_1(p^\ast, p^\theta)/\hat{L}(D_\alpha)$. This normalised quantity  should be compared with $W_1(p^\ast,p^\theta)$.
\end{remark}

\begin{figure}
    \centering
    \includegraphics[scale=.05]{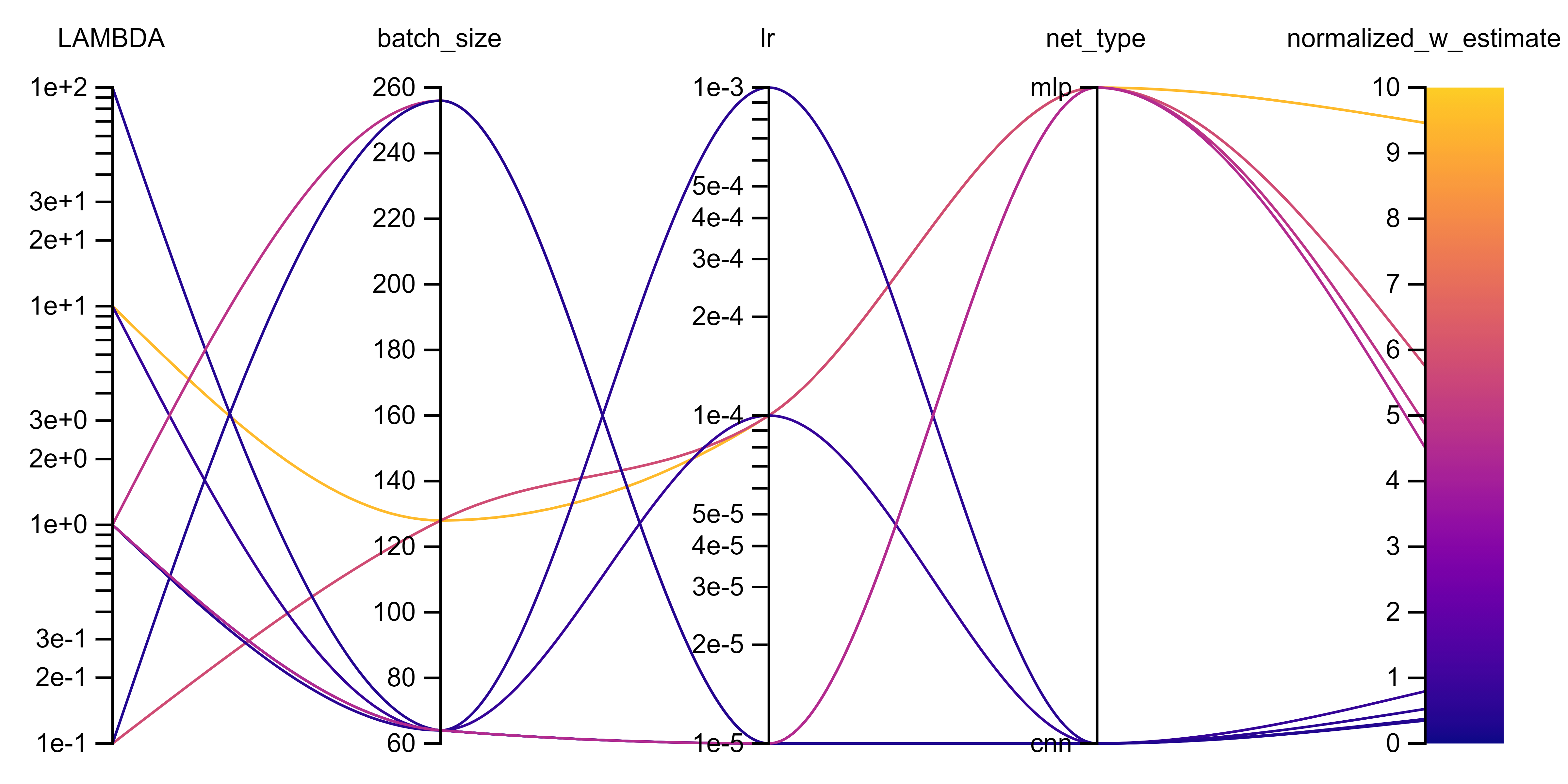}
    \caption{Estimated Wasserstein distance for different hyper-parameter configurations. The correct value is $W_1(p^\ast,p^\theta) = 41.21$. (Hyperparameter optimisation and plot done using \cite{wandb}).}
    \label{fig:W_haparams}
\end{figure}

\begin{figure}
    \includegraphics[scale=.5]{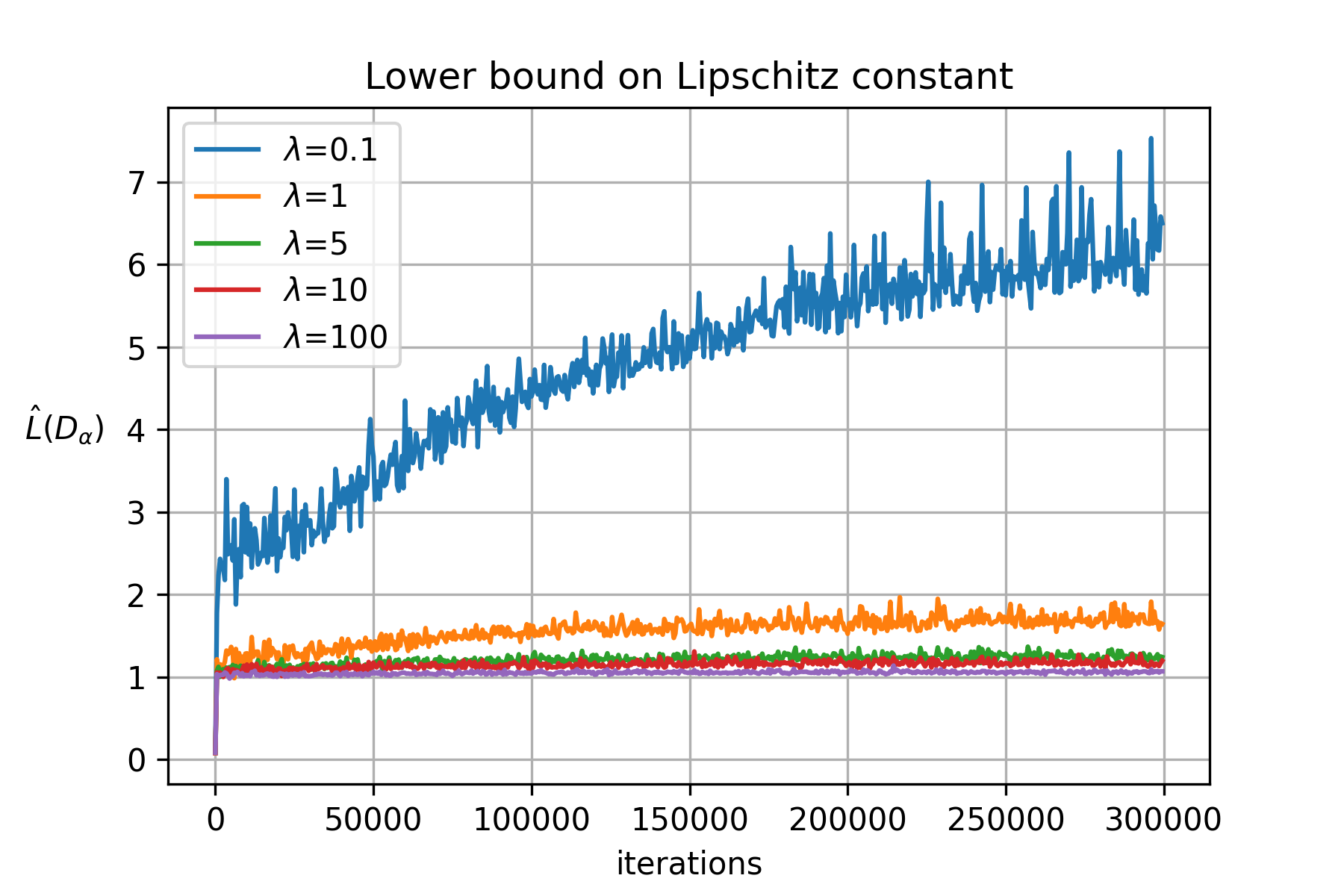}
    \includegraphics[scale=.5]{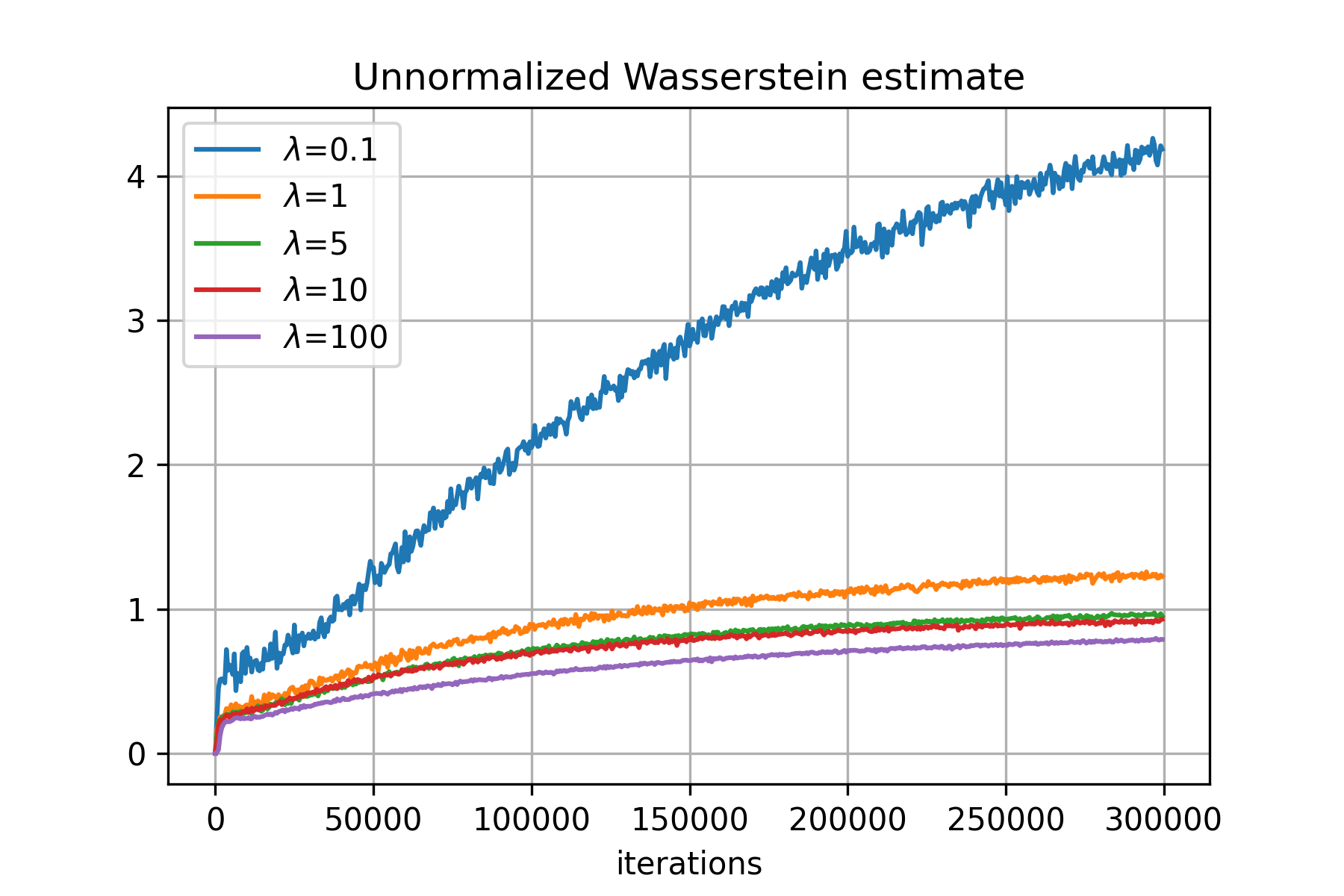}
    \includegraphics[scale=.5]{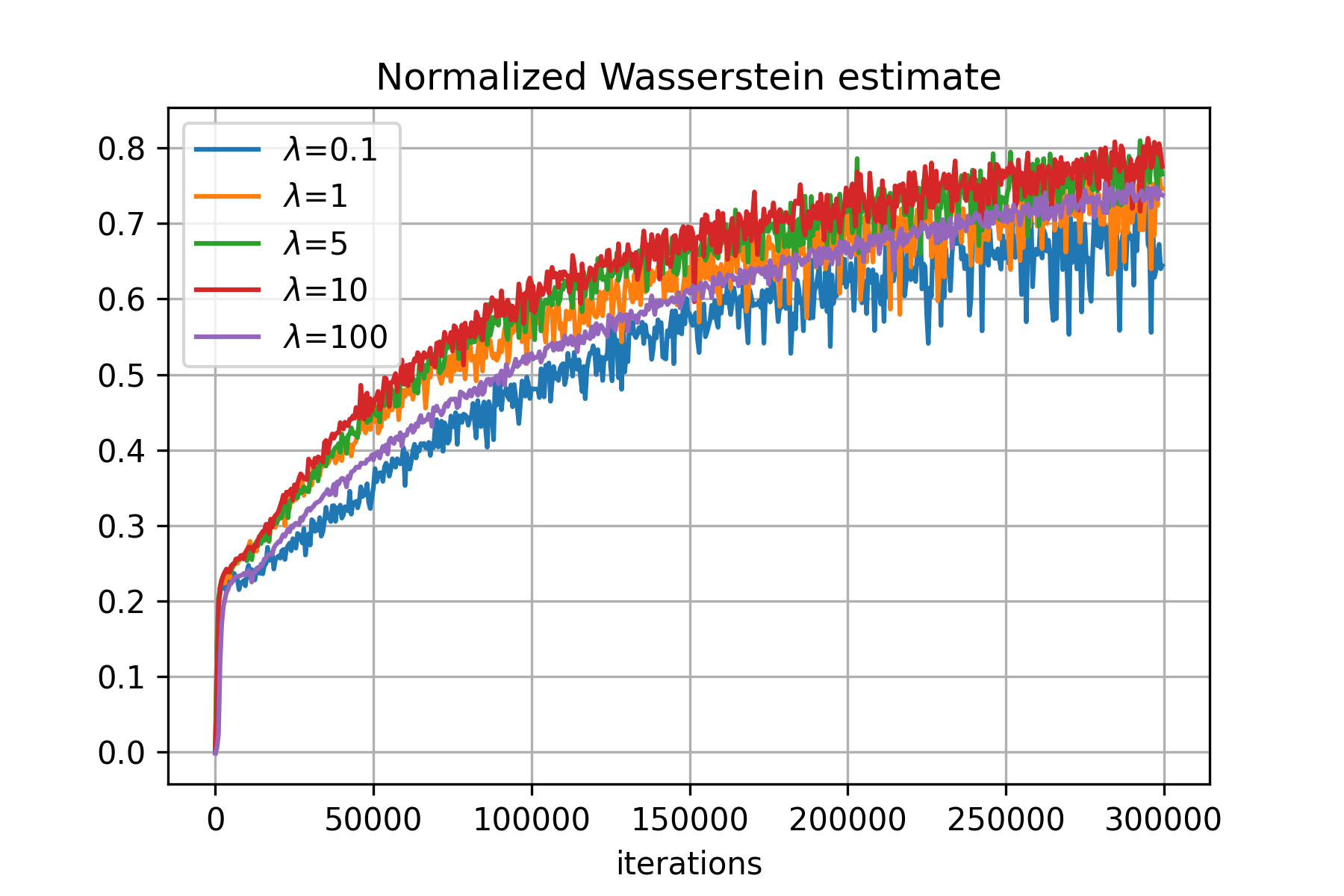}
    \caption{Estimated Wasserstein distance for different values of $\lambda$ ($\lambda=10$ is recommended in \cite{WGAN-GP}). All WGAN-GP approximations are very far from the correct value of $W_1(p^\ast,p^\theta) = 41.21$.}
    \label{fig:W_dist_lambda}
\end{figure}

In the experiments in Figures \ref{fig:W_haparams} and \ref{fig:W_dist_lambda}, we use Algorithm~\ref{alg:oracleapprox}. We find that normalised Wasserstein estimate is very far from the actual $W_1(p^\ast,p^\theta)$. Since $W_1(p^\ast,p^\theta)=41.21$, the normalised Wasserstein estimate $W^D_1(p^\ast,p^\theta)/\hat{L}(D_\alpha)$ is one order of magnitude smaller than $W_1(p^\ast,p^\theta)$. 
In the experiments in Figure \ref{fig:W_haparams} we explored a  range of hyperparameters ($\lambda$, batch size, learning rate, network architecture). For the experiments visualized in Figure \ref{fig:W_dist_lambda} we use the hyperparameters recommended in the original WGAN-GP paper \cite{WGAN-GP} except for the parameter $\lambda$ which we allowed to vary. This allows us to explore how the value of $\lambda$ influences the Lipschitz constraint and how it impacts the quality of Wasserstein estimation.


We emphasise that the task of estimating $W_1(p^\ast,p^\theta)$ in Algorithm \ref{alg:oracleapprox} is easier than estimating $W_1(p^\ast,p^\theta)$ during WGAN training. Distributions $p^\ast$ and $p^\theta$ are static, while $p^\theta$ changes after each $N_D$ updates to the discriminator $D_\alpha$ in Algorithm \ref{alg:WGAN-GP}. During WGAN training typically $N_D \leq 10$, while we allowed $D_\alpha$ to be trained for $N=300K$ iterations on the same pair of distributions. The fact that even this simpler task cannot be accomplished implies that the approximation of the Wasserstein distance during WGAN training is unrealistic with WGAN-GP algorithm \ref{alg:WGAN-GP}.

We  point out that we are  conservative in our approach to normalize the Wasserstein estimate since
\begin{gather*}
    \frac{W^D_1(p^\ast,p^\theta)}{\norm{D_\alpha}_L}  \leq \frac{W^D_1(p^\ast,p^\theta)}{\hat{L}(D_\alpha)} \ll W_1(p^\ast,p^\theta),
\end{gather*}
where the first inequality follows from $\hat{L}(D_\alpha) \leq \norm{D_\alpha}_L$, and the second inequality is supported by Figure \ref{fig:W_dist_lambda}.

We conclude that $\loss_G(\theta) \approx W^\ast(p^\ast, p^\theta)$ is not achieved in WGAN-GP, so the loss of WGAN-GP is not an accurate approximation of Wasserstein distance.

\subsubsection{Approximation during training in low dimensions}

In this section, we investigate whether  $\loss_G(\theta) \approx W^\ast(p^\ast, p^\theta)$ can be achieved for the special case of low dimensions where sample complexity issues (discussed in detail in Section \ref{sec:finite_sample_approx}) can be neglected. In low dimensions, we can approximate $W_1( p^\ast,p^\theta)$ accurately by $W_1(p^\ast_n,p_n^\theta)$ for a sufficiently large number of samples $n$ (we used $n=1000$ for this experiment). Since $p^\ast_n,p_n^\theta$ are finite measures, we can determine $W_1(p^\ast_n,p_n^\theta)$ by solving the linear program and we can check how close $\loss_G(\theta)$ is to $W_1(p^\ast,p^\theta)$ during a WGAN training.  

\begin{figure}
    \centering
    \includegraphics[scale=.33]{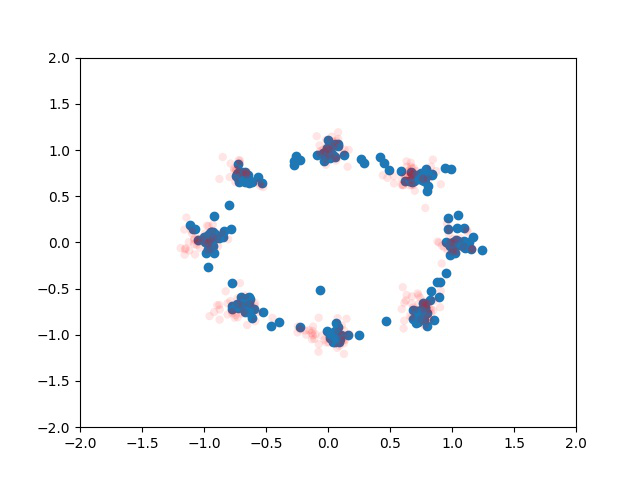}
    \caption{Blue dots are samples from $p^\theta$ and red dots are samples from $p^\ast$. }
    \label{fig:8_gauss_approx_1}
    \includegraphics[scale=.55]{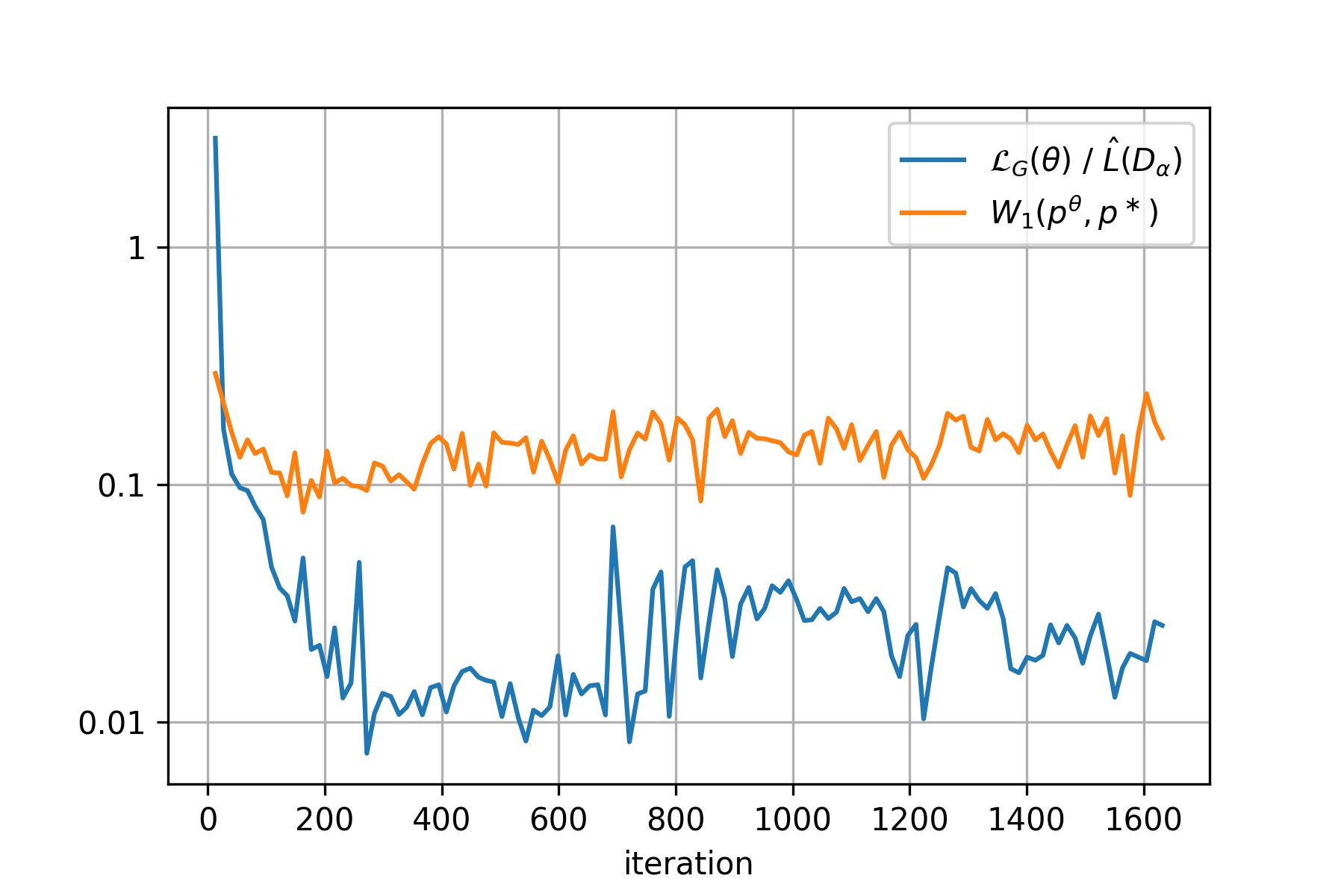}
    \caption{Normalized loss and true Wasserstein distance in log scale. At the end of the training the Wasserstein distance is 6.92 times higher than the normalized loss.}
    \label{fig:8_gauss_approx_2}
\end{figure}

We conduct an experiment where we fit the WGAN-GP in Algorithm \ref{alg:WGAN-GP} to an 8-mode Gaussian mixture and track $\loss_G(\theta), W_1(p^\ast,p^\theta)$ at each iteration. As discussed in Remark \ref{rem:lipschitznormalisation}  we normalize the loss by $\hat{L}(D_\alpha)$ as the Lipschitz constraints in WGAN-GP are only approximated. In Figure \ref{fig:8_gauss_approx_1}, we show $250$ samples from $p^\ast$ and $p^\theta$. The associated Wasserstein distance $W_1( p^\ast,p^\theta)$ and the normalised Wasserstein estimate $\loss_G(\theta)/\hat{L}(D_\alpha)$ obtained with Algorithm \ref{alg:WGAN-GP}  are shown in Figure \ref{fig:8_gauss_approx_2}. 
 We observe that the normalised loss is an order of magnitude smaller the $W_1(p^\ast,p^\theta)$ as in the Experiments in Section \ref{sec:approxfixed}. Notice that any sensible positive loss function will be close to zero. Again we conclude that even in a simple two dimensional case $\loss_G(\theta) \approx W_1( p^\ast,p^\theta)$ is not achieved. 

\subsection{Approximation of the batch estimator}
\label{sec:approximation_of_batch_estimator}
In this section we examine whether
\begin{gather*}
    \loss_G(\theta) \approx \hat{W_1}(p^\ast, p^\theta).
\end{gather*}
We  consider two algorithms: WGAN-GP \cite{WGAN-GP} and c-transform WGAN \cite{Mallasto}. We reproduce the experiment of \cite{Mallasto} in a higher resolution setup and we use an improved architecture. For experiments in this section, we use the architecture based on StyleGAN \cite{StyleGAN} and CelebA data set \cite{celeba}.

As in \cite{Mallasto}, we train WGAN according Algortihm \ref{alg:WGAN-GP}. At the end of each iteration of the generator loop we solve the linear program to evaluate the true Wasserstein distance $W_1(p^\ast_n,p^\theta_n)$ between the generated batch $p^\theta_n$ and the batch of the real data $p^\ast_n$. Then we compare the result with $\loss_G(\theta)$. For $c$-transform WGAN $\loss_G(\theta)$ is computed using the equation \ref{eq:lossapproxctransf}.

A detailed description of the experiment is included as Algorithm \ref{alg:Mallasto}.

\begin{algorithm2e}
   \SetAlgoLined
    \DontPrintSemicolon
     \For{$N_G$ iterations}{
        \For{$N_D$ iterations}{
        Sample a batch $p^*_n$ from $p^\ast$ \;
        Sample a batch $p^\theta_n$ from $p^\theta$ \;
        Ascent $\alpha$\ wrt. $\loss_D(\alpha)$ \;
        }
    Sample a batch $p^*_n$ from $p^\ast$ \;
    Sample a batch $p^\theta_n$ from $p^\theta$ \;
    $W_1(p^*_n,p^\theta_n) \gets $ Solution of LP for $p^*_n,p^\theta_n$\;
    Compare $W_1(p^*_n,p^\theta_n)$ with $\loss_G(\theta) = \mathcal V(D_\alpha, p^*_n, p^\theta_n)$ \;
    Descent $\theta$ wrt. $\loss_G(\theta)$
    }
\caption{Experiment 2: Mini-batch estimator during WGAN training}
\label{alg:Mallasto}
\end{algorithm2e}

\begin{figure}[htbp]
  \centering
  \includegraphics[scale=.35]{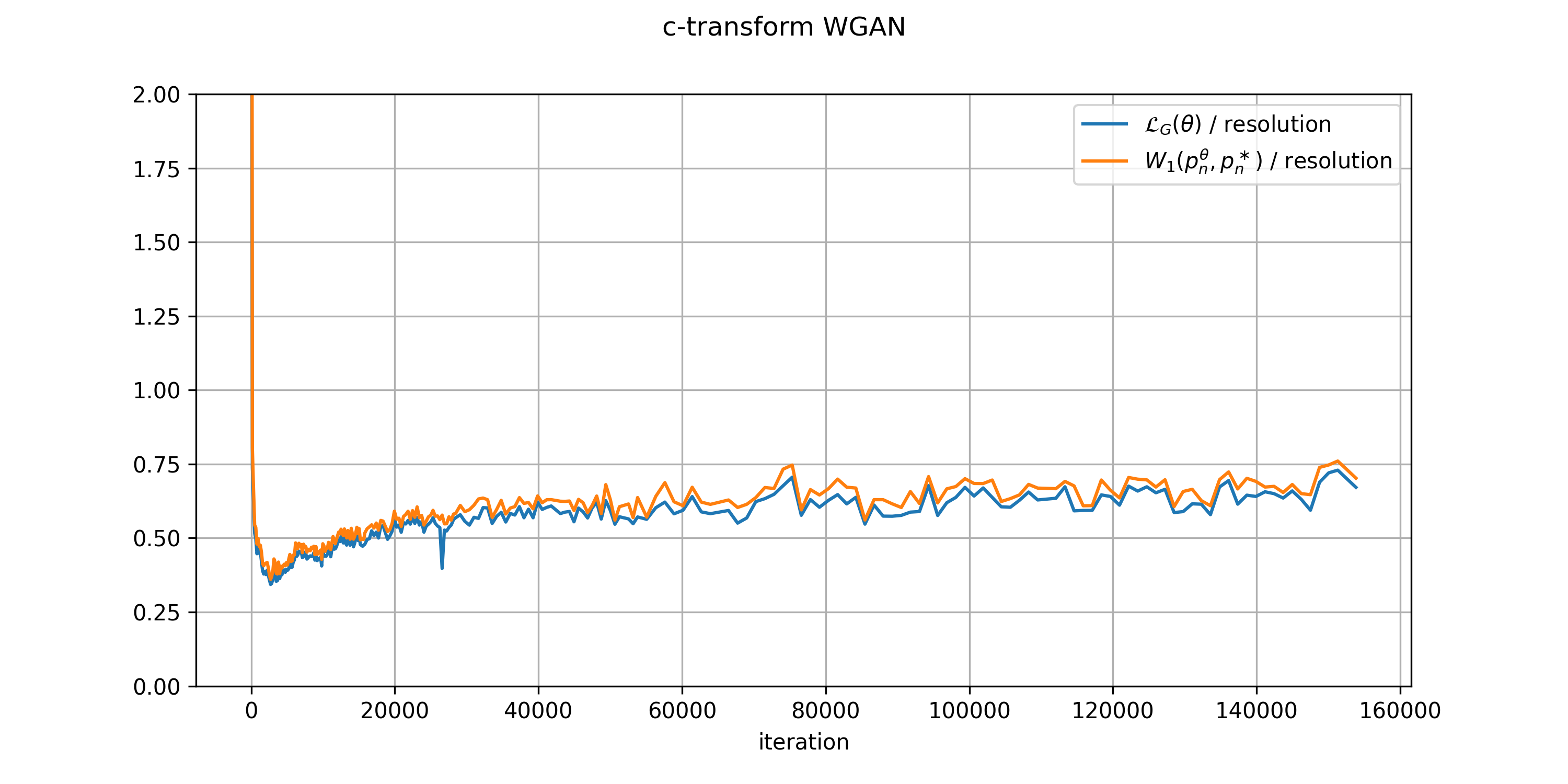}
  \includegraphics[scale=.35]{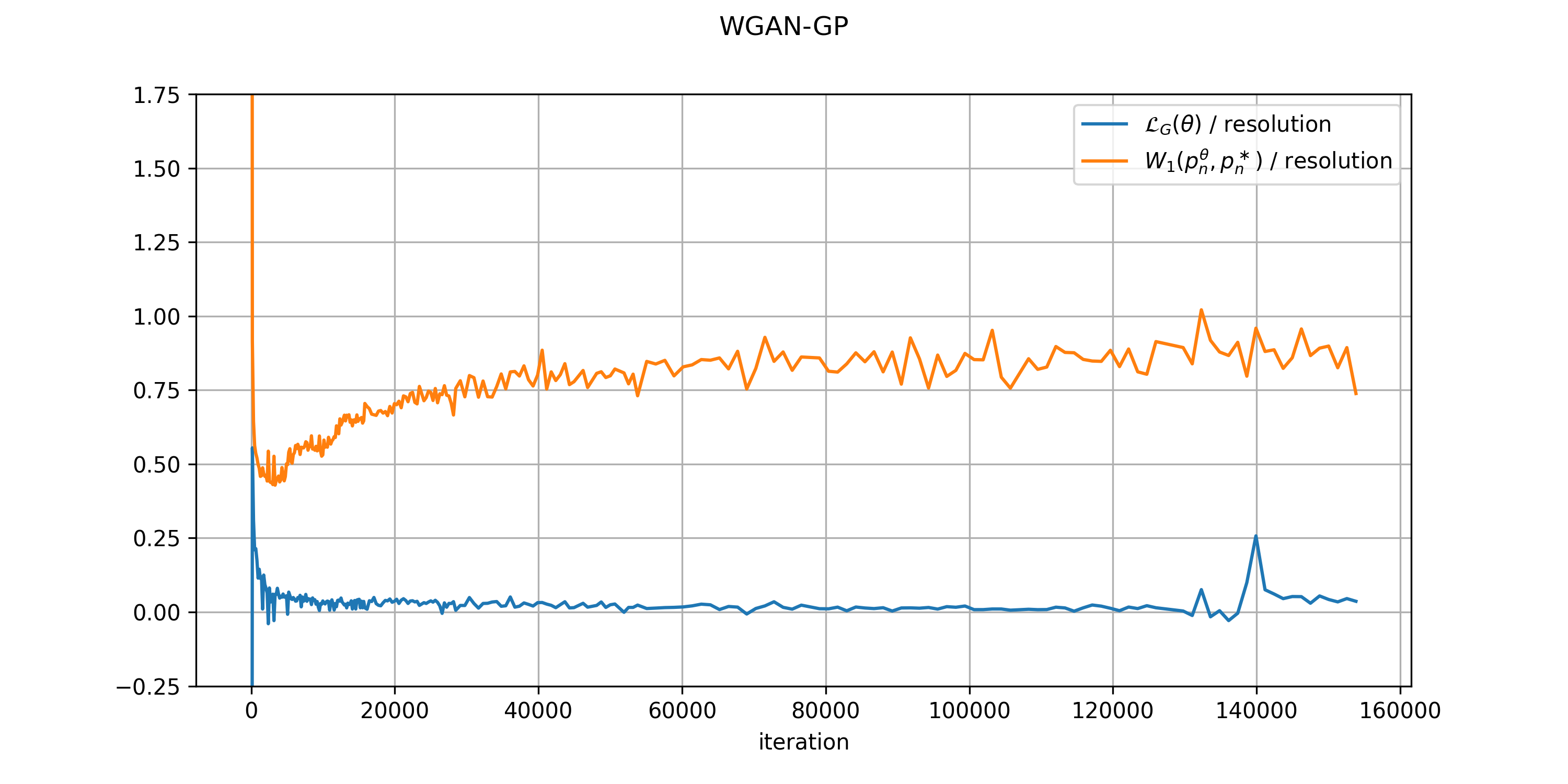}
  \caption{Plots show how accurately $W_1$ between training batches is approximated by a given method during WGAN training on the CelebA data set. 
  Based on \citep{Mallasto}.}
  \label{WGAN_Approx}
\end{figure}

\begin{figure}[htbp]
  \centering
  \includegraphics[scale=.6]{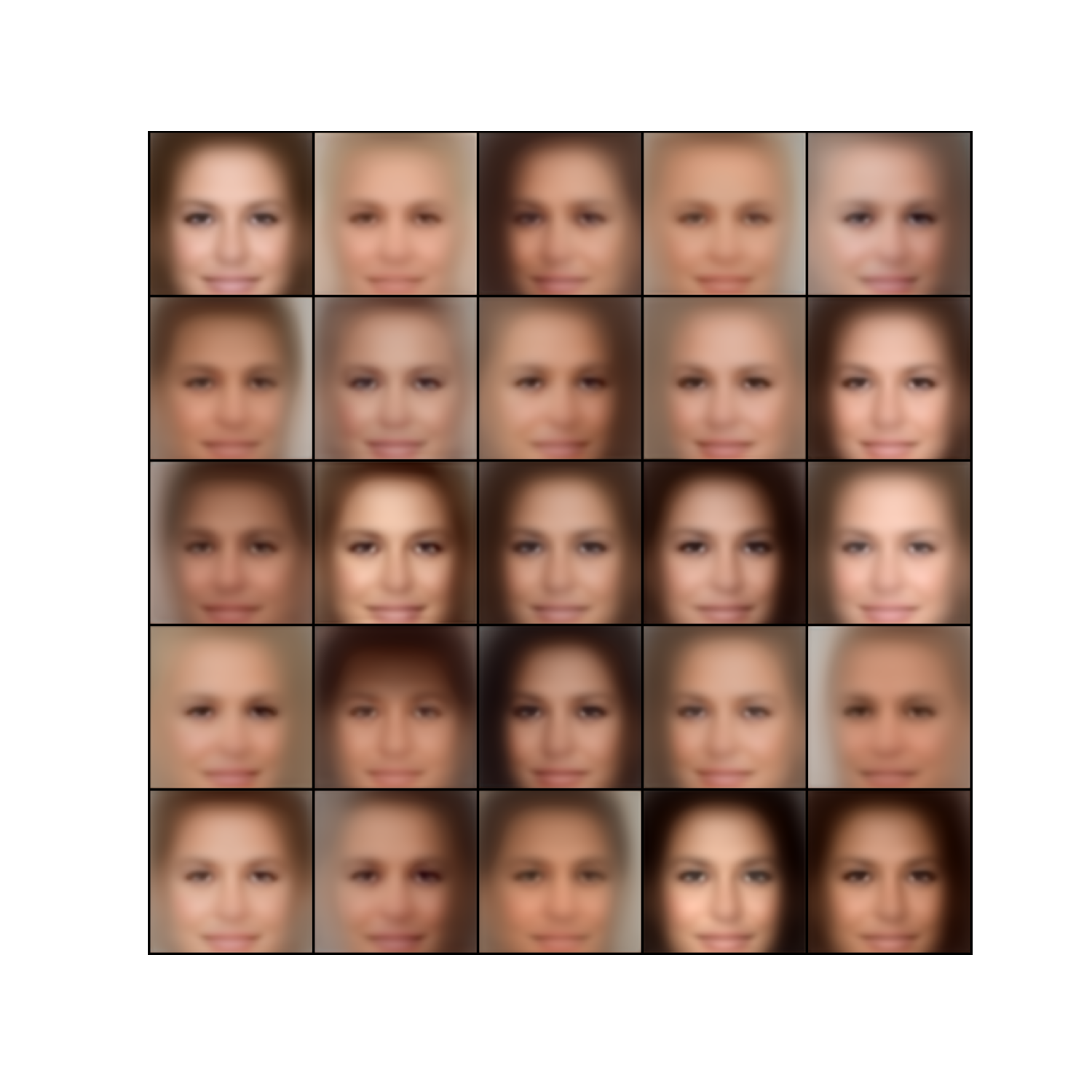}
  \includegraphics[scale=.6]{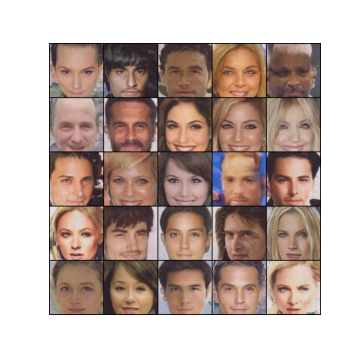}
  \caption{Samples resulting from the training with a given approximation method. $c$-transfrom on the top and gradient penalty on the bottom. Based on \citep{Mallasto}.}
  \label{WGAN_Approx2}
\end{figure}

As shown in Figure \ref{WGAN_Approx}, the gradient penalty the gradient penatly method does not provied an accurate approximation of $\hat{W}_1$. On the other hand, the $c$-transform method approximates $\hat{W_1}$ very accurately. Surprisingly, a good approximation of the batch Wasserstein distance does not correspond to a good generative performance. Figure \ref{WGAN_Approx2} shows samples obtained from training with the $c$-transform and the gradient penalty as the approximation method. The faces generated by a WGAN using the $c$-transform approximation look very blurry, while the WGAN-GP results look realistic. In particular, the images obtained with WGAN using the $c$-transform do not capture the complexity of the data set  as well as WGAN-GP, despite achieving a better approximation of the Wasserstein distance. Moreover the loss function of WGAN-GP $\loss(G)$ doesn't decrease despite the samples getting better with more training. This is because the loss of WGAN-GP reflects how well the generator performs compared to the discriminator, not the Wasserstein distance.

So we are left with a puzzling question: \textit{Why does a better approximation of the batch Wasserstein distance result in a worse generative performance?}

In next sections, we  examine possible explanations based on sample complexity, biased gradients and connections of $W_1$ to the $L_2$-norm and clustering.

\section{Finite sample approximation of $W_1$}\label{sec:finite_sample_approx}

In the following, we analyse  problems arising from the fact that we use finite data and minibatch-based optimisation to estimate the Wasserstein distance between high dimensional distributions.

\subsection{Sample complexity of Wasserstein distance estimators} \label{sec:sample_complexity_of_W_estimators}
Recall that the oracle estimator is defined as 
\begin{gather*}
    {W_1^\ast}(p_n^\ast,p^\theta_n) =  \mathbb{E}_{x \sim p_n^\ast}[f^\ast(x)] - \mathbb{E}_{x \sim p^\theta_n} [f^\ast(x)], 
\end{gather*}
 where $f^* \in \argmax_{\substack{\norm{f}_L \leq 1}} \left(\mathbb{E}_{x \sim p^\ast}[f(x)] - \mathbb{E}_{x \sim p^\theta} [f(x)] \right).$ \\

In the oracle estimator, the only effect of finite samples is the Monte Carlo approximation of the expectations which has a convergence rate of $O\big(\frac{1}{\sqrt{n}}\big)$ when $n$ samples are considered for $p^\ast_n,p^\theta_n$. The oracle estimator assumes that we have  access to an oracle which  provides us with  Kantorovich's potential $f^\ast$ between $p^\ast$ and $p^\theta$. Therefore, the true sample complexity is moved to the oracle and the above convergence rate is misleading. An efficient oracle does not exist and in practice, one  needs huge number of samples to be able to accurately estimate $f^\ast$. 

For the batch estimator
\begin{gather*}
    \hat{W}_1(p_n^\ast,p^\theta_n) =  \max_{\substack{\norm{f}_L}\leq 1}\left( \mathbb{E}_{x \sim p_n^\ast}[f(x)] - \mathbb{E}_{x \sim p^\theta_n} [f(x)] \right),    
\end{gather*}
we have  $\hat{W}_1(p_n^\ast,p^\theta_n)= {W}_1(p_n^\ast,p^\theta_n)$ and 
the sample complexity is well known. As shown in \citep{WeedBach}, for $d$-dimensional data, the expected error of the estimation of the Wasserstein distance decreases as $O(n^{-1/d})$, i.e. 
\begin{align*}
\mathbb{E}_{\substack{p_n^\ast \sim \mathcal P^\ast_n, \\ p_n^\theta \sim \mathcal P^\theta_n}}[|W_1(p_n^\ast,p_n^\theta) - W_1(p^\ast,p^\theta)|] = O(n^{-1/d}),
\end{align*}
where $\mathcal P_n^\ast$ and $\mathcal P_n^\theta$ denote the sets of all empirical measures of $n$ samples drawn from $p^\ast$ and $p^\theta$, respectively.
This decay rate is very slow in high dimensions, and hence, even if the optimal discriminator between $p_n^\ast$ and $p_n^\theta$ is learned perfectly, the loss function of WGAN is very far away from the actual Wasserstein distance.

In the following sections, we argue that sample complexity issues render the oracle estimator unrealistic  and the mini-batch estimator useless.
\subsection{Empirical study of sample complexity issues}\label{sec:empirical_study_of_sample_complexity_issues}
In this empirical study, we illustrate that the sample size necessary for an accurate Wasserstein approximation is infeasible in the setting of high dimensional deep generative modelling. To this aim, we examine  the difference between $W_1(p_n^\ast, \tilde{p}_n^\ast)$ and $W_1(p^\ast,p^\ast)=0$ numerically where $p^\ast$  is a standard Gaussian measure in $d$ dimensions, and $p_n^\ast,\tilde{p}_n^\ast$ are empirical measures of $n$ samples drawn from $p^\ast$. Note that $W_1(p_n^\ast, \tilde{p}_n^\ast)$ decreases to 0 as $n\to \infty$ and the convergence is  $O(n^{-1/d})$. 

 The sample Wasserstein distance concentrates very well around its expectation \cite{WeedBach}. Therefore, the behaviour of the random variable $W_1(p_n^\ast,\tilde{p}_n^\ast)$ can be understood by examining $$\mathbb{E}_{\substack{p_n^\ast \sim \mathcal P_n^\ast, \tilde{p}_n^\ast \sim \mathcal P_n^\ast}}[W_1(p_n^\ast,\tilde{p}_n^\ast)],$$
 where  $\mathcal P_n^\ast$ denotes the set of all empirical measures of $n$ samples drawn from $p^\ast$. %This justifies why we examine $\mathbb{E}_{\substack{p_n \sim P, \tilde{p}_n \sim P}}[W_1(p_n,\tilde{p}_n)]$ in the following.

According to the \textit{manifold hypothesis} \cite{manifold} the distribution $p^\ast$ which we want to learn is concentrated around a lower dimensional manifold $\mathcal{M}$. According to results in \cite{WeedBach}  the dimension of $\mathcal{M}$, known as the \textit{intrinsic dimension} of $p^\ast$,  is relevant for the sample complexity of the Wasserstein distance and may be smaller than the dimension of the ambient Euclidean space. The dimension of the manifold modeled by a GAN is at most the dimension of its latent space $\latent$ \cite{TrainingGANs}. Therefore, ideally we want to set the dimension of $\latent$ to match the dimension of $\mathcal{M}$, although, when training GANs in practice, the dimension of $\latent$ is often set to $100$ or more \cite{dcgan, StyleGAN}. 

\begin{figure}[htbp]
    \centering
    \includegraphics[scale=.5]{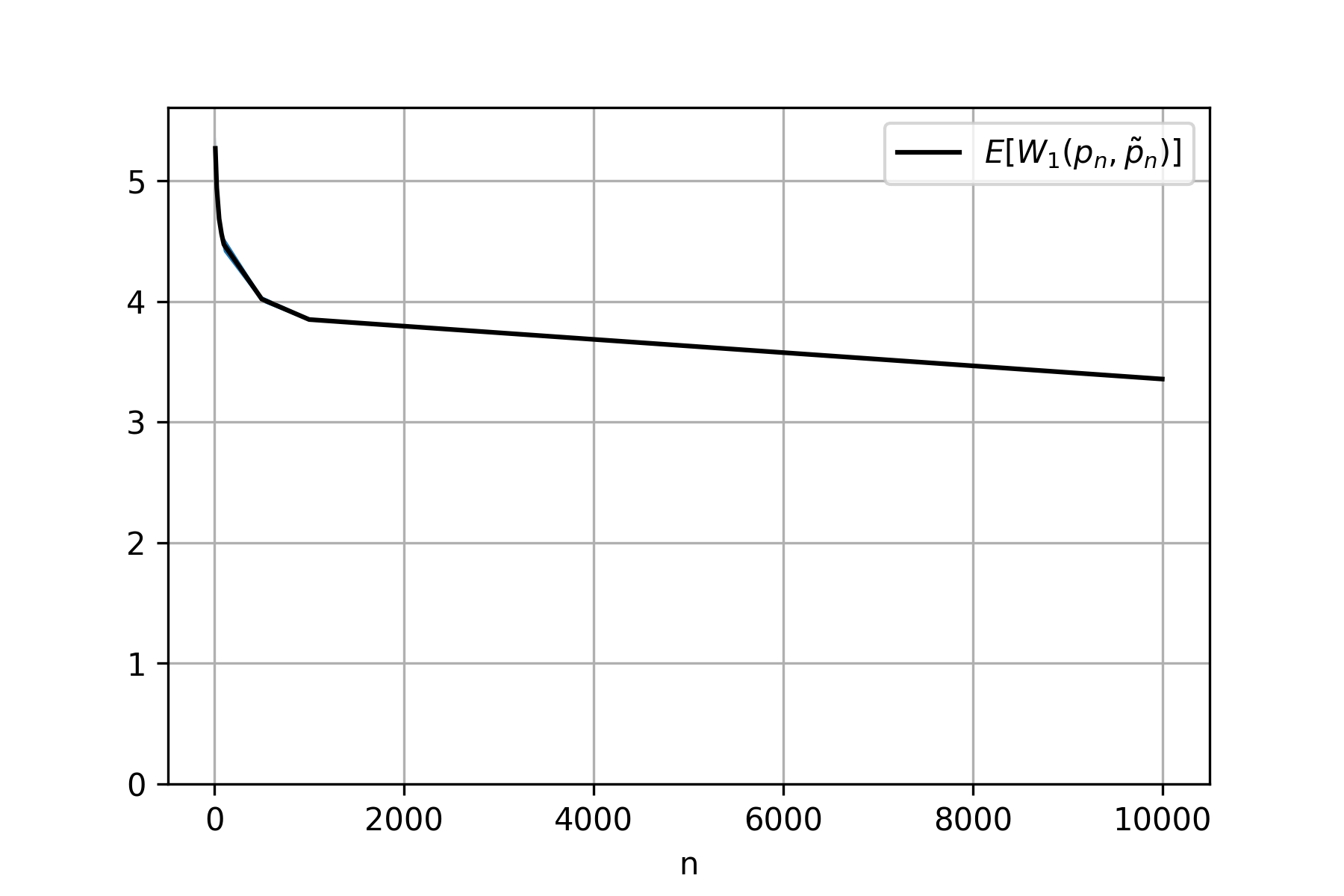}
    \includegraphics[scale=.5]{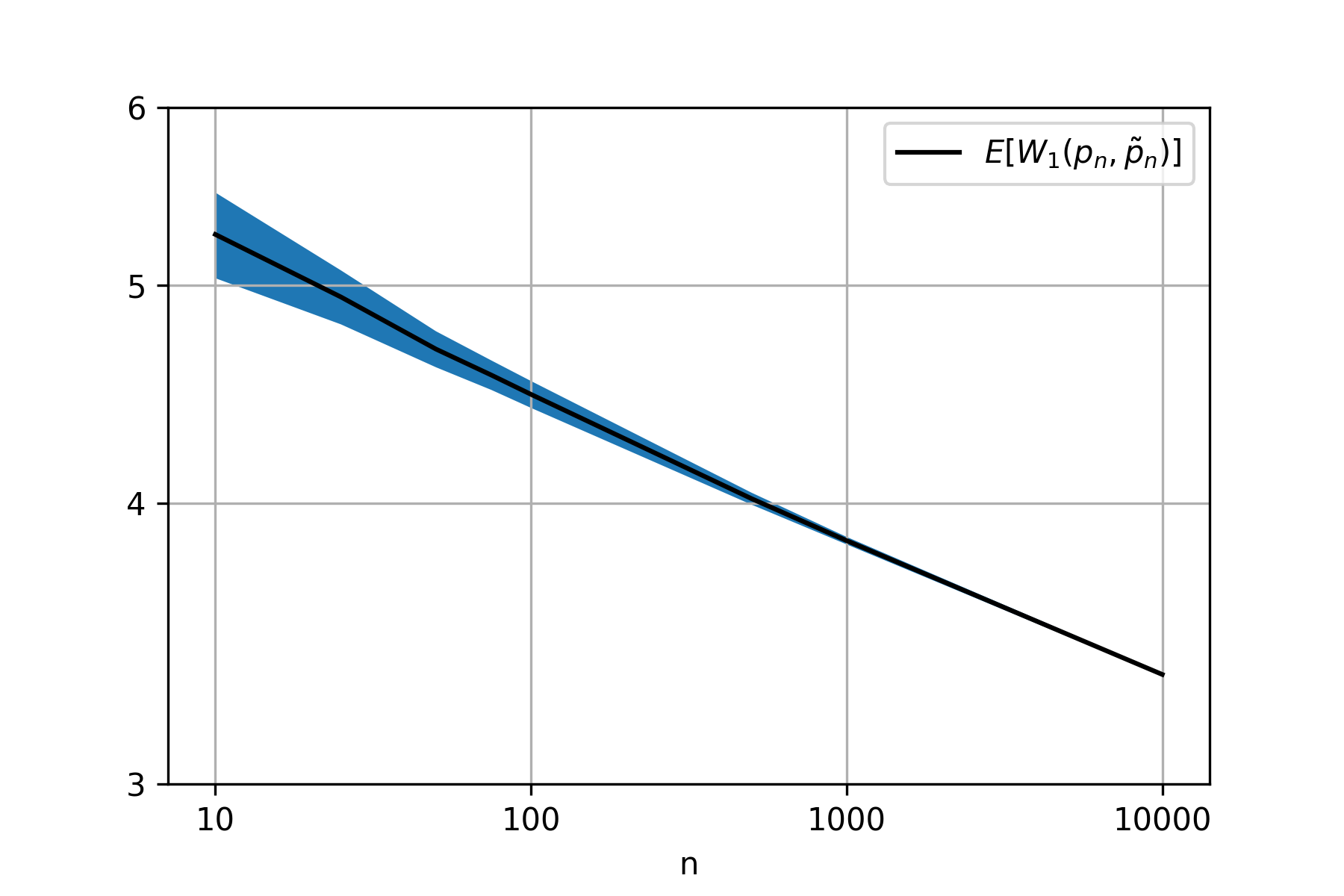}
    \caption{The figure shows the averaged (estimated using 100 repetitions of the experiment, blue bar shows standard deviation)  $W_1$ between two random samples from standard Gaussian distribution in dimension 20 as a function of the sample size. Second plot is in log-log scale. The distance converges to zero extremely slowly rendering accurate Wasserstein estimation impossible.}
    \label{fig:hyerpcube_experiment}
\end{figure}

Recent research on the intrinsic dimension suggests that benchmark data sets like CIFAR-10 and CelebA could have an intrinsic dimension of around $20$ \cite{pope2021intrinsic}. To illustrate that the sample size necessary for an accurate Wasserstein approximation is infeasible for high-dimensional deep generative modelling, we  examine the case of $d=20$ for the Wasserstein distance between two random samples from the standard Gaussian distribution in Figure \ref{fig:hyerpcube_experiment}. In our experiments,  we sample $N=300$ pairs of batches $p_n^\ast$, $\hat{p}_n^\ast$ for  $n\in\{10, 25, 50 ,75, 1000, 10000\}$ from $p^\ast$ and calculate the corresponding Wasserstein distance $W_1(p_n^\ast,\tilde{p}_n^\ast)$. Then we use a standard Monte Carlo estimator to approximate $\mathbb{E}_{\substack{p_n^\ast \sim \mathcal P_n^\ast, \tilde{p}_n^\ast \sim \mathcal P_n^\ast}}[W_1(p_n^\ast,\tilde{p}_n^\ast)]$. Even for very large batches (up to 10,000) and for a simple Gaussian distribution, the estimation of the true Wasserstein distance (=0) is extremely bad.

%$$\mathbb{E}_{\substack{p_n^\ast \sim \mathcal P_n^\ast, \tilde{p}_n^\ast \sim \mathcal P_n^\ast}}[W_1(p_n^\ast,\tilde{p}_n^\ast)] \approx \frac{1}{N}\sum W_1(p_n^\ast,\tilde{p}_n^\ast)$$.

\begin{figure}[htbp]
    \centering
    \includegraphics[scale=.5]{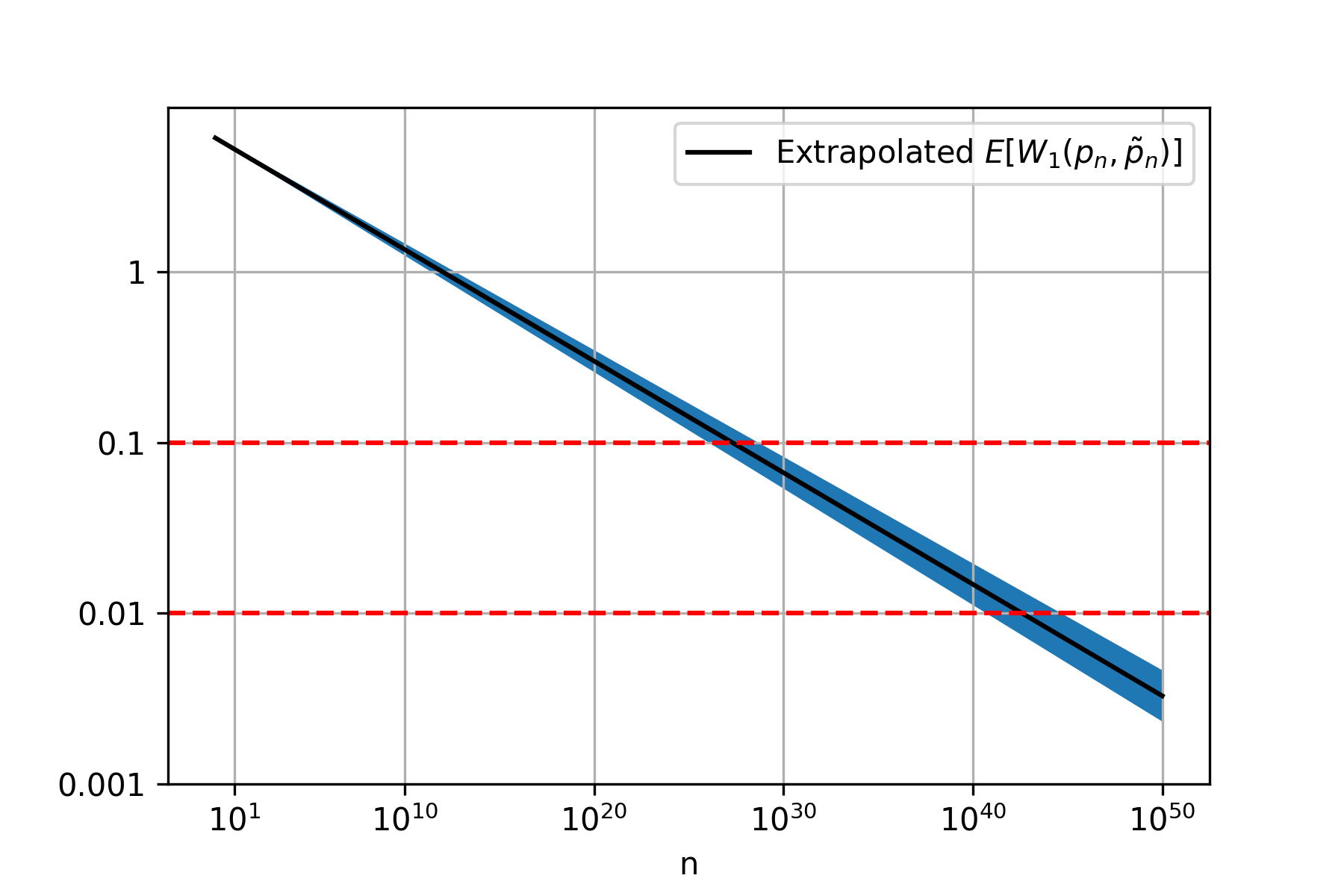}
    \caption{Extrapolation in log-log scale. Blue bar is the $95\%$ confidence interval.}
    \label{fig:hyerpcube_experiment_extrapolation}
\end{figure}

Using the fact that in log-log space the relationship is linear, we fit a least squares line and extrapolate for larger values of $n$ in Figure \ref{fig:hyerpcube_experiment_extrapolation}. In this way we establish that in order to bring the approximation error to $0.1$ one would need over $10^{20}$ samples, which is much larger than any conceivable data set. In order to bring the error down to $0.01$ one would need over $10^{40}$ samples.

\subsection{False minima of the batch Wasserstein distance}
\label{sec:why_batch_estimator_poor_loss}

Given a target  distribution $p^\ast$ and generated distribution $p^\theta$ with parameter $\theta$, the difference between the true Wasserstein distance $W_1( p^*,p^\theta)$ and its sample estimate $W_1( p^*_n,p^\theta_n)$ may cause the existence of `false' global optima, i.e.\ $\min_\theta  \E_{\substack{ p_n^\ast \sim \mathcal P_n^\ast,p_n^\theta \sim \mathcal P_n^\theta,}}[W_1(p^*_n,p^\theta_n)]$  may be different from $\min_\theta W_1(p^*,p^\theta)$, where $\mathcal P_n^\ast$ and $\mathcal P_n^\theta$ denote the sets of all empirical measures of $n$ samples drawn from $p^\ast$ and $p^\theta$, respectively.

An example of this phenomenon has already been pointed out in \citep{Bellemare}. The authors demonstrate that false global minima may appear when one tries to learn a Bernoulli measure  by minimising the batch Wasserstein distance. For the target Bernoulli measure $p^\ast$ and the generated Bernoulli measure $p^\theta$ with parameter $\theta\in(0,1)$, they show that the sample estimate of the Wasserstein gradient  $\nabla_\theta W_1(p^*_n,p^\theta_n)$ is a biased estimator of $\nabla_\theta W_1(p^*,p^\theta)$. These estimation errors can strongly affect the training via stochastic gradient descent (SGD) or SGD-based algorithms like Adam. Their results are summarised in the following theorem:
\begin{theorem} \citep{Bellemare}\label{thm:falsemin}
    %Let $p_n^\ast$ be the empirical distribution derived from $n$ i.i.d.\ samples from a Bernoulli distribution $p^\ast$. Then 
    \begin{enumerate}
        \item \emph{Non-vanishing mini-max bias of the sample gradient.} For any $n \geq 1$ there exists a pair of Bernoulli distributions $p^\ast$, $p^\theta$ such that
            \begin{gather*}
                | \ \mathbb{E}{\substack{p_n^\ast \sim \mathcal P_n^\ast}}[\nabla_\theta W_1(p_n^\ast, p^\theta)] - \nabla_\theta W_1(p^\ast, p^\theta) \ | \geq 2e^{-2}.
            \end{gather*}
        \item \emph{Wrong minimum of the batch Wasserstein loss.} For Bernoulli measures $p^\ast,p^\theta$, the minimum of the expected sample loss $$\bar{\theta} =\argmin_\theta \mathbb{E}_{\substack{p_n \sim \mathcal P_n^\ast}}[W_1(p_n^\ast, p^\theta)]$$ is in general different from the minimum of the true Wasserstein loss $\theta^\ast = \argmin_\theta W_1(p^\ast, p^\theta)$.
    \end{enumerate}
\end{theorem}

While Theorem \ref{thm:falsemin} shows that minima of the distributional and the batch Wasserstein distance may not coincide, we investigate the existence of `false' minima of the batch Wasserstein distance further in the context of WGAN training and show empirically that false minima can  appear while learning synthetic (e.g.\ Gaussian) and benchmark distributions (e.g.\ CelebA \citep{celeba}). For this, we consider certain fixed batches (a 
`real batch' $\tilde{p}^\ast_n$, a  `mean batch' $p^\mu$ and a `geometric $k$-medians batch' $p^{k-\text{gm}}$). We show empirically that in sufficiently high dimensions the expected Wasserstein distance between these batches and $p_n^\ast \sim \mathcal P_n^\ast$ is largest for $\mathbb{E}_{\substack{p_n^\ast \sim \mathcal P_n^\ast}}[W_1(p_n^\ast, \tilde{p}_n^\ast)]$, even though $p_n^\ast,\tilde{p}_n^\ast$ are both empirical measures with samples drawn from $p^\ast$ and $W_1(p^\ast, p^\ast)=0$. This is achieved by approximating the expectation  with the standard Monte Carlo estimator using 100 sample batches from $p^\ast$.  We conclude that a `mean batch' or a `geometric $k$-medians batch' provide false minima in the case of the CelebA data set.

\begin{figure}
     \centering
     \begin{subfigure}{.5\textwidth}
         \centering
         \includegraphics[width=.5\textwidth]{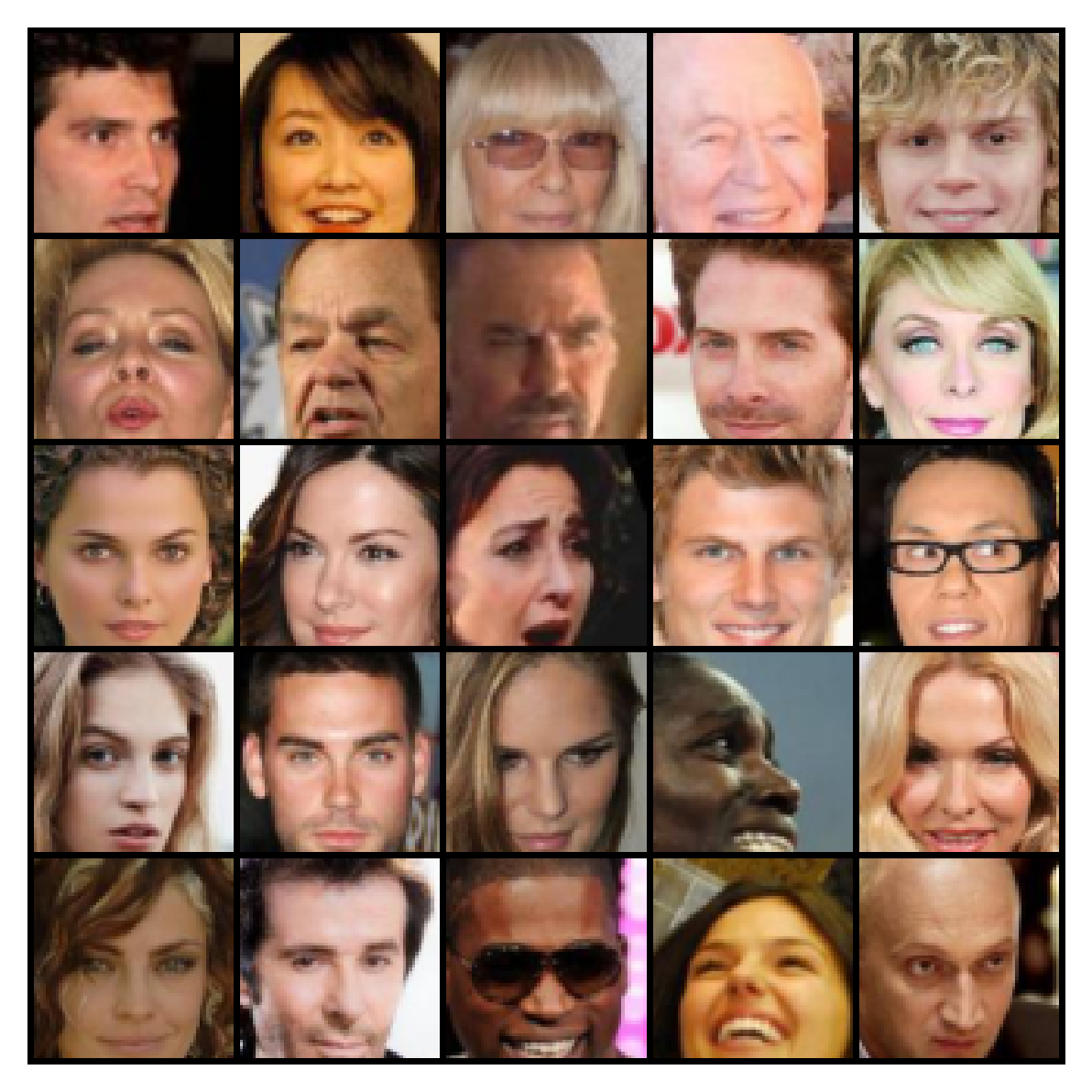}
         \caption{$\mathbb{E}_{\substack{p_n^\ast \sim \mathcal P_n^\ast}}[W_1(p_n^\ast, \tilde{p}_n^\ast) ]=50.67$}
         \label{fig:random_batch}
     \end{subfigure}
     \hfill
     \begin{subfigure}{.5\textwidth}
         \centering
         \includegraphics[width=.5\textwidth]{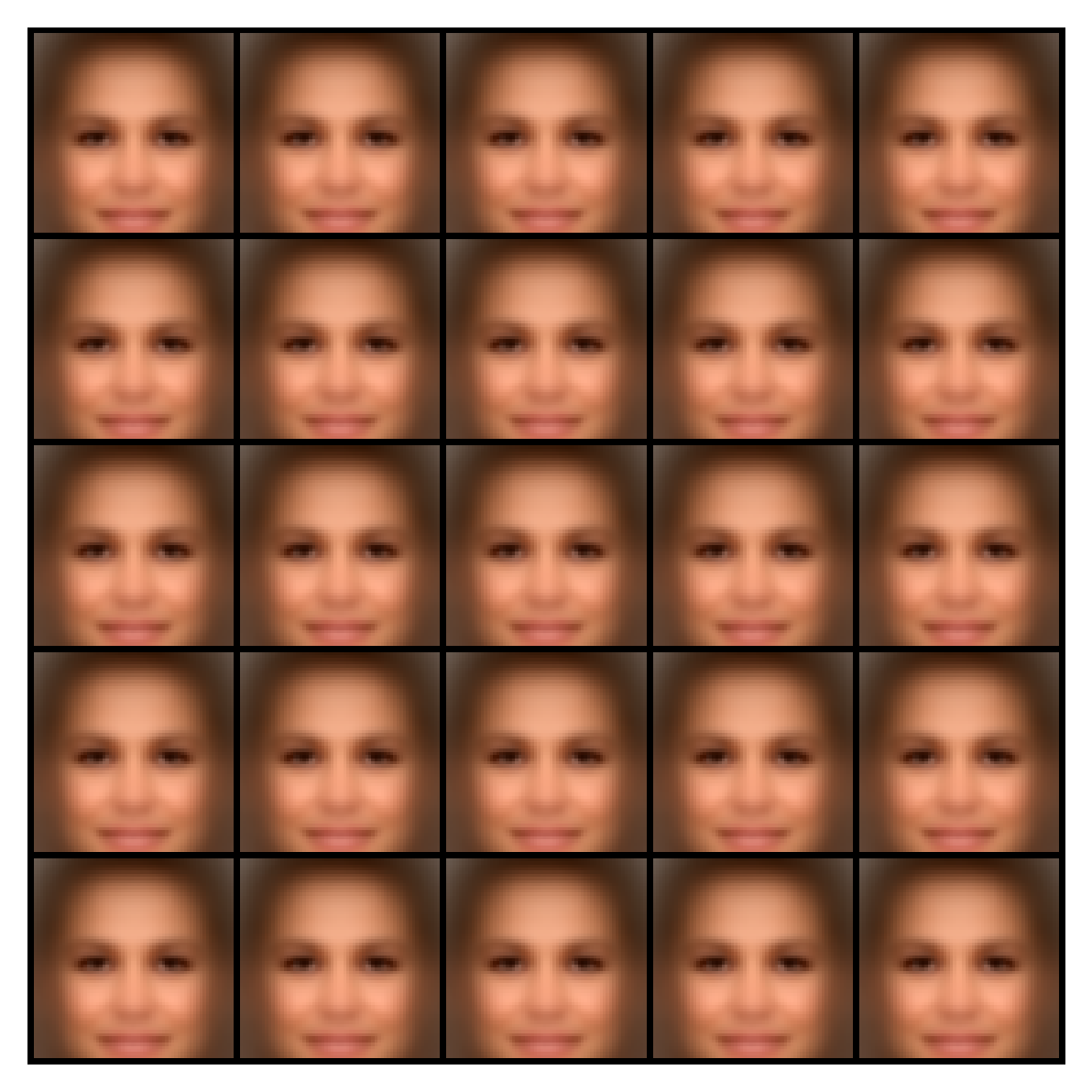}
         \caption{$\mathbb{E}_{\substack{p_n^\ast \sim \mathcal P_n^\ast}}[W_1(p_n^\ast, p^\mu) ]=47.91$}
         \label{fig:mean_batch}
     \end{subfigure}
     \hfill
     \begin{subfigure}{.5\textwidth}
         \centering
         \includegraphics[width=.5\textwidth]{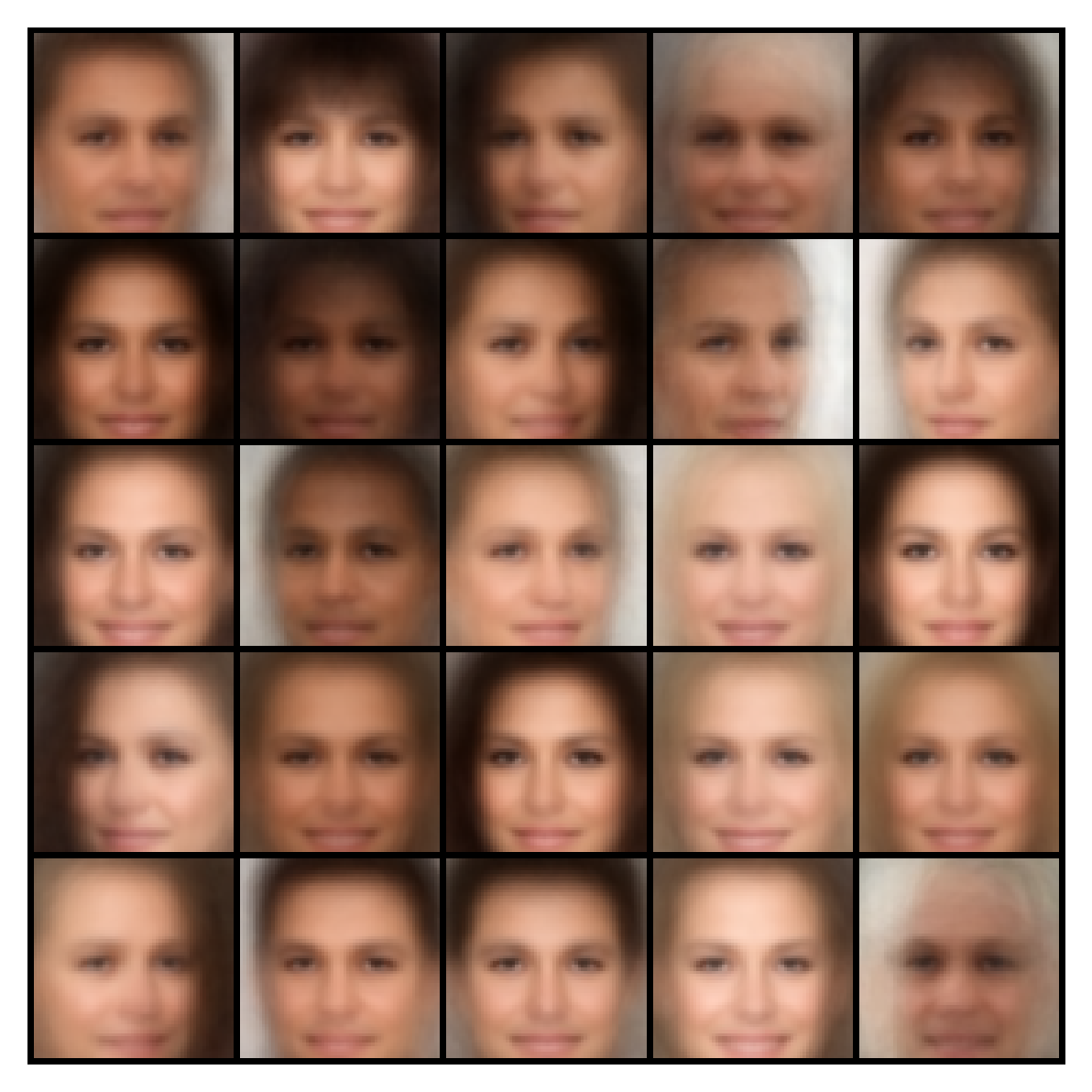}
         \caption{$\mathbb{E}_{\substack{p_n^\ast \sim \mathcal P_n^\ast}}[W_1(p_n^\ast, p^\text{k-gm}) ]=39.44$}
         \label{fig:gkm_batch}
     \end{subfigure}
        \caption{Three batches and their respective batchwise Wasserstein distance (for CelebA). We observe that a batch $\tilde{p}_n$ of real faces (Fig. \ref{fig:random_batch}) has higher batchwise $W_1$-distance than undesirably simple generated batches: In particular, both a batch $p^\mu$ of the 'average face' (Fig.\ \ref{fig:mean_batch}), as well as a batch $p^\text{k-gm}$  of the $k$ centroids of geometric $k$-medians clustering (Fig. \ref{fig:gkm_batch}) over the data set (for $k=n$) yield far lower batchwise $W_1$-distance than real data.}
        \label{fig:batches}
\end{figure}

The results of the above experiment are visualised in the Figure \ref{fig:batches} for CelebA.  We show that the expected batch Wasserstein distance between two samples from the target distribution $\mathbb{E}_{\substack{p_n^\ast \sim \mathcal P_n^\ast}}[W_1(p_n^\ast, \tilde{p}_n^\ast)]$ is greater than the expected batch Wasserstein distance between a sample from the target distribution and a sample consisting of repeated means $\mathbb{E}_{\substack{p_n^\ast \sim \mathcal P_n^\ast}}[W_1(p_n^\ast, p^\mu)]$ (or geometric $k$-medians). Therefore, a \emph{perfect} generator producing samples from the target distribution $p^\ast$ would have (on average) a greater loss than a generator which learned a Dirac distribution concentrated at the mean of $p^*$. Hence, the batch Wasserstein distance can push the generator towards false minima making it an undesirable loss function.

\begin{figure}
    \centering
    \includegraphics[scale=.6]{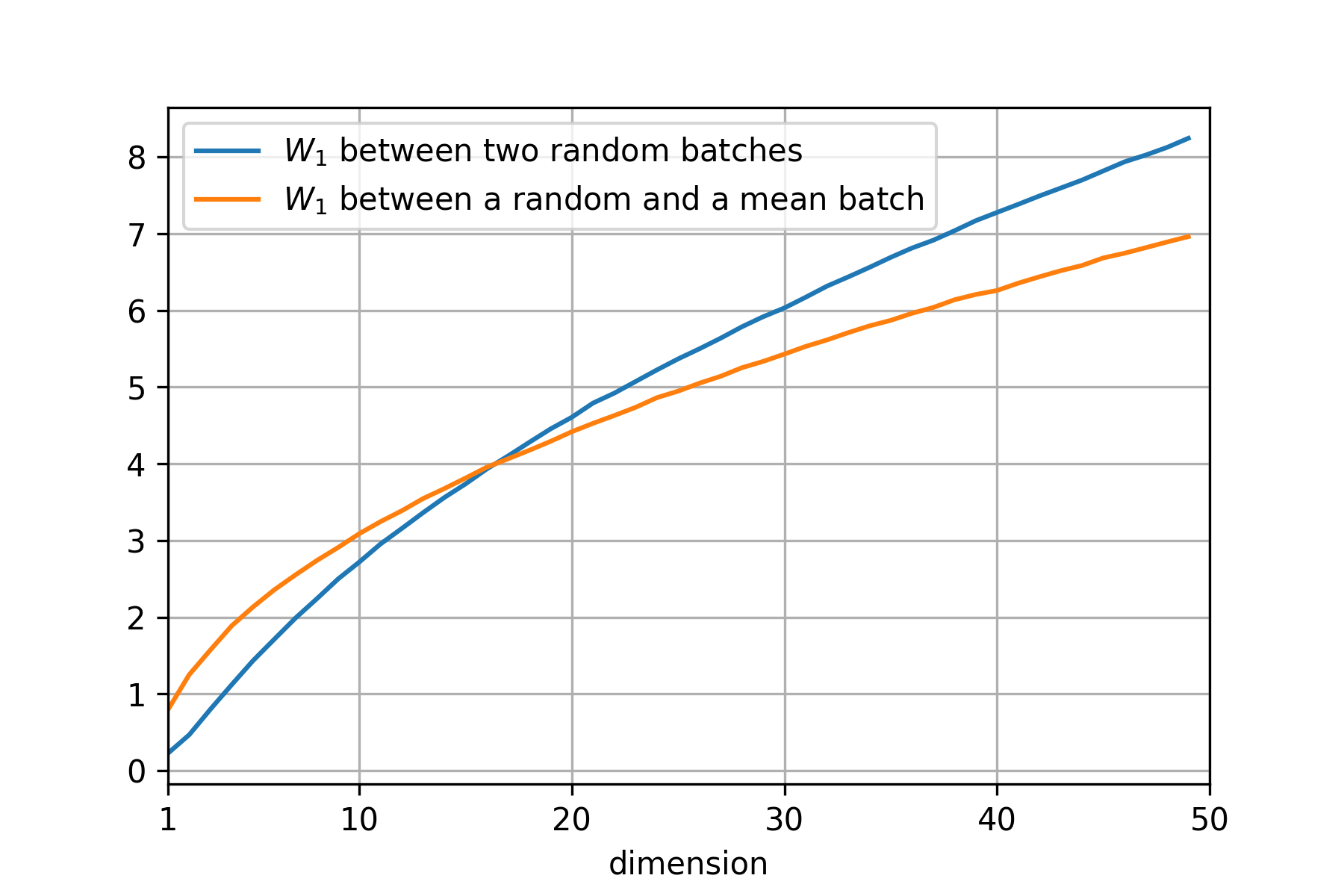}
    \caption{In sufficiently high dimension, a distribution concentrated on the mean has (on average) a smaller batch Wasserstein distance from a random sample of $p^*$ than another random sample of $p^*$. Here, we can see this in the case of the Gaussian measure.}
    \label{gaussian}
\end{figure}

Next we compare $\mathbb{E}_{\substack{p_n^\ast \sim \mathcal P_n^\ast}}[W_1(p_n^\ast, \tilde{p}_n^\ast)]$ and $\mathbb{E}_{\substack{p_n^\ast \sim \mathcal P_n^\ast}}[W_1(p_n^\ast, p^\mu)]$ for the case where $p^\ast$ is the standard Gaussian distribution and $p^\mu$ is a Dirac distribution concentrated at its mean, as a function of dimension $d$. As before,  the expectation is approximated with the standard Monte Carlo estimator using 100 sample batches from $p^\ast$. The results are visualized on Figure \ref{gaussian}. Again we observe that in sufficiently high dimension ($d > 15$) we have $\mathbb{E}_{\substack{p_n^\ast \sim \mathcal P_n^\ast}}[W_1(p_n^\ast, p^\mu)] < \mathbb{E}_{\substack{p_n^\ast \sim \mathcal P_n^\ast}}[W_1(p_n^\ast, \tilde{p}_n^\ast)]$. Therefore as in the case of the CelebA dataset, a generator producing samples from the target Gaussian distribution $p^\ast$ has (on average) a greater batch Wasserstein loss than a generator which learned a Dirac distribution concentrated at the mean of $p^*$ in high dimensions.

%\LK{add short conclusion why the batch estimator is a poor loss function}

\subsection{Connection to clustering}\label{sec:connection_to_clustering}

\begin{figure}[htbp]
    \centering
    \includegraphics[scale=.40]{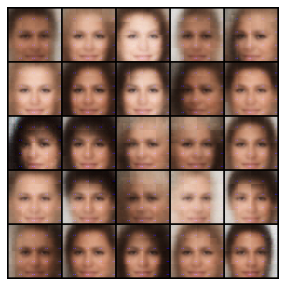}
    \includegraphics[scale=.40]{gkm_clusters.png}
    \caption{On the left: Output of WGAN trained with c-transform \citep{Mallasto}. On the right: centroids produced by geometric $k$-medians clustering.}
    \label{ctrans-kmean}
\end{figure}

As we pointed out in Section  \ref{sec:approximation_of_batch_estimator}, training a WGAN using the $c$-transform results in a better approximation of the batch Wasserstein distance, but achieves much worse generative performance than WGAN-GP. Moreover, as one can observe in  Figure \ref{ctrans-kmean}, the output of the $c$-transform WGAN looks very similar to geometric $k$-medians clustering.

As discussed in Section \ref{sec:empirical_study_of_sample_complexity_issues}, clustering may provide samples of low batch Wasserstein distance with low visual fidelity. \citep{canas} show a connection between $k$-medians clustering and the Wasserstein-2 distance. Here, we extend these results to the Wasserstein-1 distance $W_1$, which is the basis for Wasserstein GANs. Our result establishes a connection between to $W_1$ and \emph{geometric $k$-medians clustering}. 

The well-known $k$-means clustering determines  cluster centroids so that the sum of the \emph{squared} euclidean distances of the cluster elements to their respective centroid is minimal. Suppose we are given a finite set $\mathcal{X} \subseteq \mathbb{R}^d$. Given an arbitrary set $S=\{m_i, i=1,\ldots,k\} \subseteq \mathbb{R}^d$, let  $S_i = \{ x \in \mathcal X : m_i = \argmin_{m \in S} \norm{m - x}_2 \}$, i.e.\ $S_i$ consists of points in $\mathcal{X}$ for which the closest point in $S$ is $m_i$. K-means is the solution to the  optimisation problem
\begin{gather*}
    \argmin_{\substack{S \subseteq \mathbb{R}^d : |S|=k}} \sum_{i=1}^k \sum_{x \in \substack{S_i}} \norm{m_i - x}_2^2.
\end{gather*}

If we replace the squared euclidean distance by the (unsquared) euclidean distances, we obtain geometric $k$-medians clustering, given by
\begin{gather*}
    \argmin_{S \subseteq \mathbb{R}^d : |S|=k} \sum_{i=1}^k \sum_{x \in \substack{S_i}} \norm{m_i - x}_2.
\end{gather*}

Note that for an arbitrary finite subset $S$ of an euclidean space $$\argmin_m \sum\limits_{x \in S} \|m-x\|_2$$  is called the \emph{geometric median} of $S$. Unlike in the case of squared distances (where the arithmetic mean provides the respective minimiser), iterative algorithms have to be used for computing the geometric median of a set. In this work, we use Weiszfeld's algorithm \citep{weiszfeld1937point} to calculate the geometric median.

As for $k$-means clustering, finding the true, globally optimal clustering is infeasible in practice due to the NP-hardness of the problem. We employ the standard LLoyd's algorithm \citep{lloyd} for clustering, which is a popular choice for $k$-means clustering. In contrast to $k$-means clustering (where centroids are computed via the arithmetic mean), we use  Weiszfeld's algorithm to compute the centroids, which lie in the geometric medians of the individual clusters. We initialise our clustering method with the $k$-means clustering method in \texttt{scikit-learn} \citep{sklearn}, as we noticed that  the geometric medians lie quite close to the arithmetic means in practice. We used 100 different random initialisations for the $k$-means clustering to find the best loss for the geometric $k$-medians clustering.

Let $p^\text{k-gm}$ be an empirical measure formed by geometric $k$-medians centroids. Similarly to the results for the Wasserstein-2 distance \citep{canas}, we show that in fact
    \begin{gather*}
      p^\text{k-gm} = \argmin_{|\text{supp}(p)| = k} W_1(p, p^*)    
    \end{gather*}
    in Theorem \ref{th:w1-kmedians}.
This means that the distribution concentrated at the centroids of geometric $k$-medians has the smallest $W_1$ distance from the target distribution $p^*$ among all distributions supported on $k$ points. 

This result could shed some light on why geometric $k$-medians clustering creates undesirable minima for the batch Wasserstein distance. Training data comes in mini-batches and if we set the mini-batch size as $k$, then the batch consisting of geometric $k$-medians centroids is the mini-batch with the smallest possible loss. 

There are still some differences between this analysis and the actual WGAN training which is based on the batch Wasserstein distance. We use a sample $p^\ast_k$ from the target measure, rather than the full target measure $p^\ast$. Hence, when using batch estimator as loss, we actually minimise
\begin{gather*}
  \mathbb{E_{\substack{p_k^\ast \sim \mathcal P^\ast_k,\\ p^\theta_k \sim P^\theta_k}}}[W_1(p_k^\ast,p^\theta_k)]  
\end{gather*}
wrt.\ $\theta$.
We leave the problem of formally extending Theorem \ref{th:w1-kmedians} as a possible objective for further study. We notice that in Section \ref{sec:why_batch_estimator_poor_loss} we have already seen empirically (in the case of CelebA) that $\E[W_1(p^\text{k-gm}, p^*_k)] < \E[W_1(p^*_k, \tilde{p}^*_k)]$, where $p^*_k, \tilde{p}^*_k$ are two empirical measures, each  formed by $k$ i.i.d. samples from $p^*$.

\section{Fundamental issues of the Wasserstein distance with image data}\label{sec:fundamental_problems}

In the previous sections, we have argued several ways, in which WGANs may fail to learn the Wasserstein distance. 
\begin{itemize}
    \item In Section \ref{sec:approximation_of_oracle}, we saw that WGANs used in practice distinctly fail to approximate the true oracle discriminator $f^\ast$, i.e.\ they fail to minimise the distributional Wasserstein distance.
    \item In Section \ref{sec:approximation_of_batch_estimator}, we saw that WGANs with gradient penalties distinctly fail to approximate the batch Wasserstein distance.
    \item As elaborated in Sections \ref{sec:sample_complexity_of_W_estimators} and \ref{sec:empirical_study_of_sample_complexity_issues}, (batch) Wasserstein estimators require  larger data sets than feasible in practice in order to accurately approximate the distributional Wasserstein distance. We further saw that batch Wasserstein estimators may yield false (Section \ref{sec:why_batch_estimator_poor_loss}) or undesirable (Section \ref{sec:connection_to_clustering}) minima.
\end{itemize}
In this section, we will show that even if some of these problems are mitigated, the resulting generators will produce undesirable images. This observation is based on the fact that the euclidean metric, i.e.\ the $L_2$-distance based on pixelwise differences, is  used explicitly in the primal formulation of the Wasserstein distance in  \eqref{eq:primal_formulation} and implicitly  in its dual formulation \eqref{eq:dual_formulation}.
We make the following conjecture: The use of the euclidean metric in the definition of the Wasserstein distance \eqref{eq:primal_formulation}--\eqref{eq:dual_formulation} is fundamentally unsuited for image data in the context of generative models. 

\subsection{Mitigation of sample complexity issues}
In the following, let $W_{1, \epsilon}$ be the regularized Wasserstein distance introduced by \citep{cuturi2013sinkhorn}
\begin{align*}
    W_{1, \epsilon}(p^\ast, p^\theta) :=  \inf_{\substack{\gamma\in \Gamma(p^\ast,p^\theta)}}  &\left(\mathbb{E}_{(x,y) \sim \gamma}[\norm{x - y}] \right.\\ %\label{eq:primal_formulation} \\ 
    &\left.+ \varepsilon \ \text{KL}(\gamma \ | \ p^\ast \otimes p^\theta)\right),
\end{align*}
where the infimum is taken over all joint distributions $\gamma$ with marginals $p^\ast$ and $p^\theta$ , $\otimes$ denotes the product measure and KL is the Kullback-Leibler divergence.

Since $W_{1, \epsilon}(p, p) \not = 0$, following \citep{SinkhornGAN}, we use the  Sinkhorn divergence  $\mathcal{S}_{1,\varepsilon}$ defined as
\begin{align*}
    \mathcal{S}_{1,\varepsilon}(p^\ast, p^\theta) = &W_{1, \epsilon}(p^\ast, p^\theta) \\ &- \frac{1}{2} \left(  W_{1, \epsilon}(p^\ast, p^\ast) +  W_{1, \epsilon}(p^\theta, p^\theta) \right),
\end{align*}
which is  a normalized version of $ W_{1, \epsilon}$ satisfying $\mathcal{S}_{1,\varepsilon}(p, p) = 0$.

For $\varepsilon \to 0$, $\mathcal{S}_{1,\varepsilon}$ converges to the Wasserstein distance, whereas for $\varepsilon \to \infty$, it converges to the maximum mean discrepancy (MMD) \citep{feydy2018interpolating}. Since the MMD is insensitive to the curse of dimensionality \citep{SinkhornComplexity}, the Sinkhorn divergence can be viewed as an approximation of the Wasserstein distance, which does not suffer from sample complexity issues.  In particular, asymptotically as $\varepsilon \to \infty$, we have
$$\mathbb{E}[|\mathcal{S}_{1, \varepsilon}(p_n^\ast,p_n^\theta) - \mathcal{S}_{1,\varepsilon}(p^\ast,p^\theta)|] = O\bigg(\frac{1}{\sqrt{n}}\bigg),$$
where $p_n^\ast,p_n^\theta$ are batches of size $n$ from $p^\ast,p^\theta$, respectively.
\begin{figure}
  \centering
  \includegraphics[scale=.6]{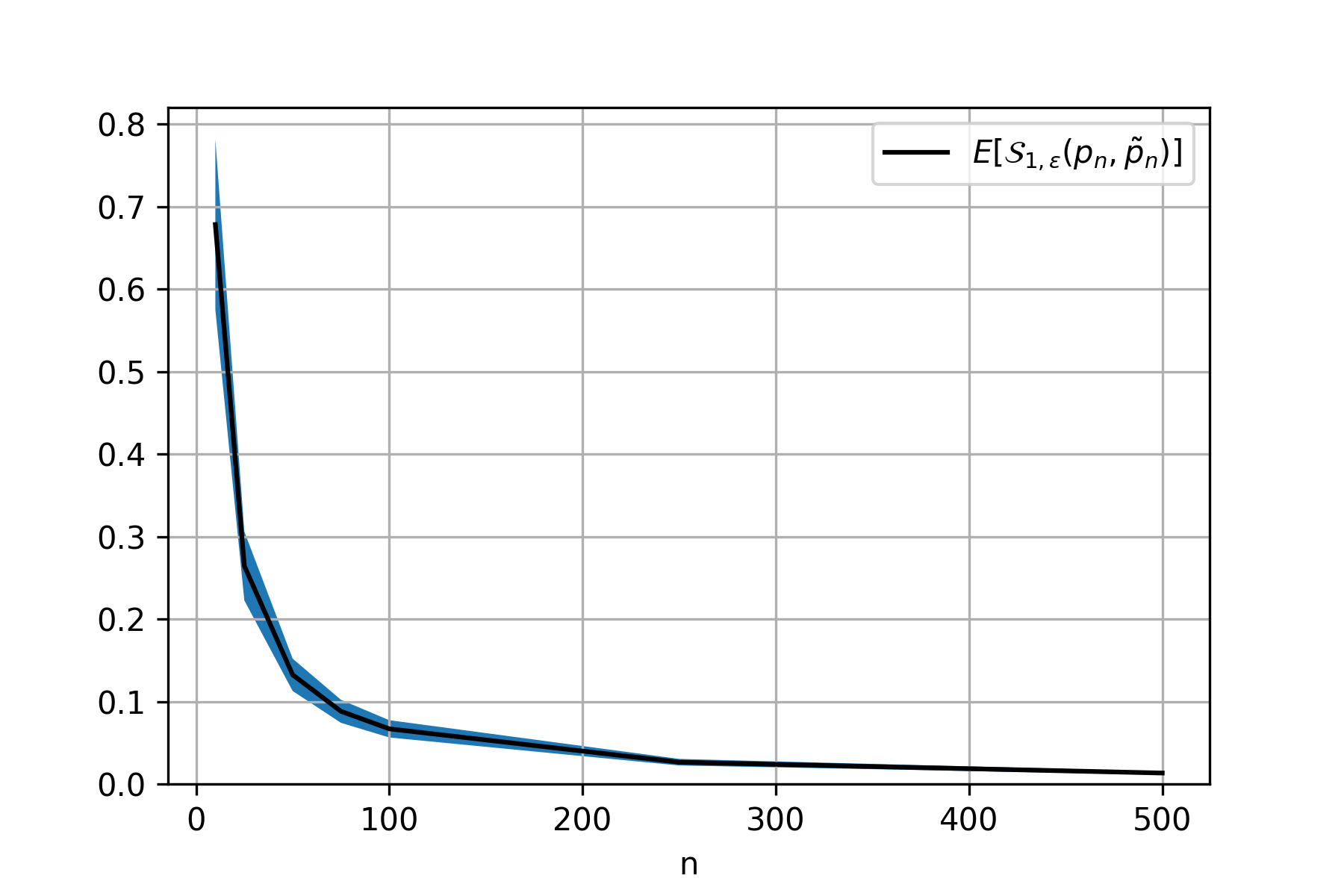}
  \caption{For high values of $\varepsilon$ (here $\epsilon=100$), the Sinkhorn divergence has a much better sample complexity than Wasserstein distance. Here the same experiment as in Figure \ref{fig:hyerpcube_experiment} was conducted. We can see that for sample size 500 the value of sample Sinkhorn divergence is aready close to 0 (contrary to batch Wasserstein distance). } 
  \label{SinkhornComplexity} 
\end{figure}

To visualize the effect on regularisation on sample complexity,we repeat the experiments from Section \ref{sec:empirical_study_of_sample_complexity_issues}. The setup is exactly the same as before, except we use Sinkhorn divergence with regularisation parameter $\varepsilon=100$ instead of the Wasserstain distance. We compute the Sinkhorn divergence between the sample batches using the Sinkhorn iterations \cite{cuturi2013sinkhorn}. The results are visualized in Figure \ref{SinkhornComplexity}. We can see that for sample size $n=500$ the value of expected Sinkhorn divergence is already close to 0 (contrary to what we have seen with Wasserstein distance in Section \ref{sec:empirical_study_of_sample_complexity_issues}).

\begin{figure}[htb]
  \centering
  \includegraphics[scale=.5]{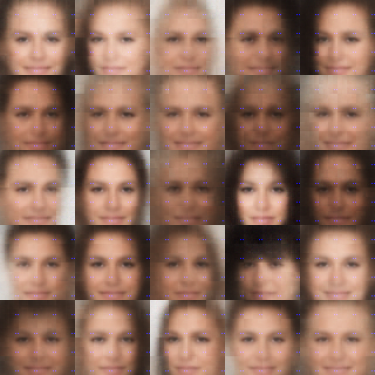}
  \caption{The output of WGAN trained using $(c,\varepsilon)$-transform with $\varepsilon=1000$.}
  \label{SinkhornGAN}
\end{figure}

In \citep{Mallasto}, a method for the estimation of batch Sinkhorn distances during GAN training was introduced, which is called the $(c, \varepsilon)$-transform.    Surprisingly, even for high values of $\varepsilon$ and thus despite a better sample complexity, the generative performance remains poor (as already noticed by \citep{Mallasto}). The generator  converges to a geometric $k$-medians-like distribution. The outputs, obtained in the  same way  as in Section \ref{sec:approximation_of_batch_estimator}, except that we replace the $c$-transform by the $(c,\epsilon)$-transform, are visualised in Figure \ref{SinkhornGAN}. This demonstrates that mitigating sample complexity issues does not solve the problems with WGANs.

\subsection{Mitigation of discriminator suboptimality issues}

If we want to use the batch Wasserstein distance or Sinkhorn divergence as a loss for a generative model,  we may take advantage of the fact that computation of Sinkhorn divergence admits automatic differentiation \cite{SinkhornGAN}. Therefore we may replace the learnable discriminator  by the true batch Sinkhorn distance and still compute the gradients using backpropagation. Moreover for small values of the regularization parameter $\varepsilon$ this will provide a faithful approximation of the Wasserstein distance \cite{feydy2018interpolating}. Such framework has been examined in \citep{learning_mb}. The authors suggest to minimise 
\begin{gather*}
  \mathbb{E_{\substack{p^\ast_n \sim P^\ast_n,\\ p^\theta_n \sim P^\theta_n }}}[\mathcal{S}_{1,\varepsilon}(p^\ast_n,p^\theta_n)]
\end{gather*}
wrt.\ $\theta$,
where the expectation is estimated using mini-batches sampled from $P^\ast_n$ and $P^\theta_n$.

We emphasise that $\mathcal{S}_{1,\varepsilon}$ is computed exactly for each mini-batch using the Sinkhorn algorithm \cite{cuturi2013sinkhorn} instead of considering the approximation of the Sinkhorn distance in the discriminator loop as in the $(c,\varepsilon)$-transform WGAN. 

We notice that this algorithm is equivalent to training a WGAN, where the discriminator optimisation step, i.e.\ the inner loop in Algorithm \ref{alg:WGAN-GP}, is replaced by the true batch Sinkhorn distance. For sufficiently low values of $\varepsilon$ we obtain a very good approximation of the optimal discriminator dynamics of batch Wasserstein GANs where the discriminator is trained to optimality for each generator update.

\begin{figure}
     \centering
     \begin{subfigure}{.5\textwidth}
         \centering
         \includegraphics[width=.5\textwidth]{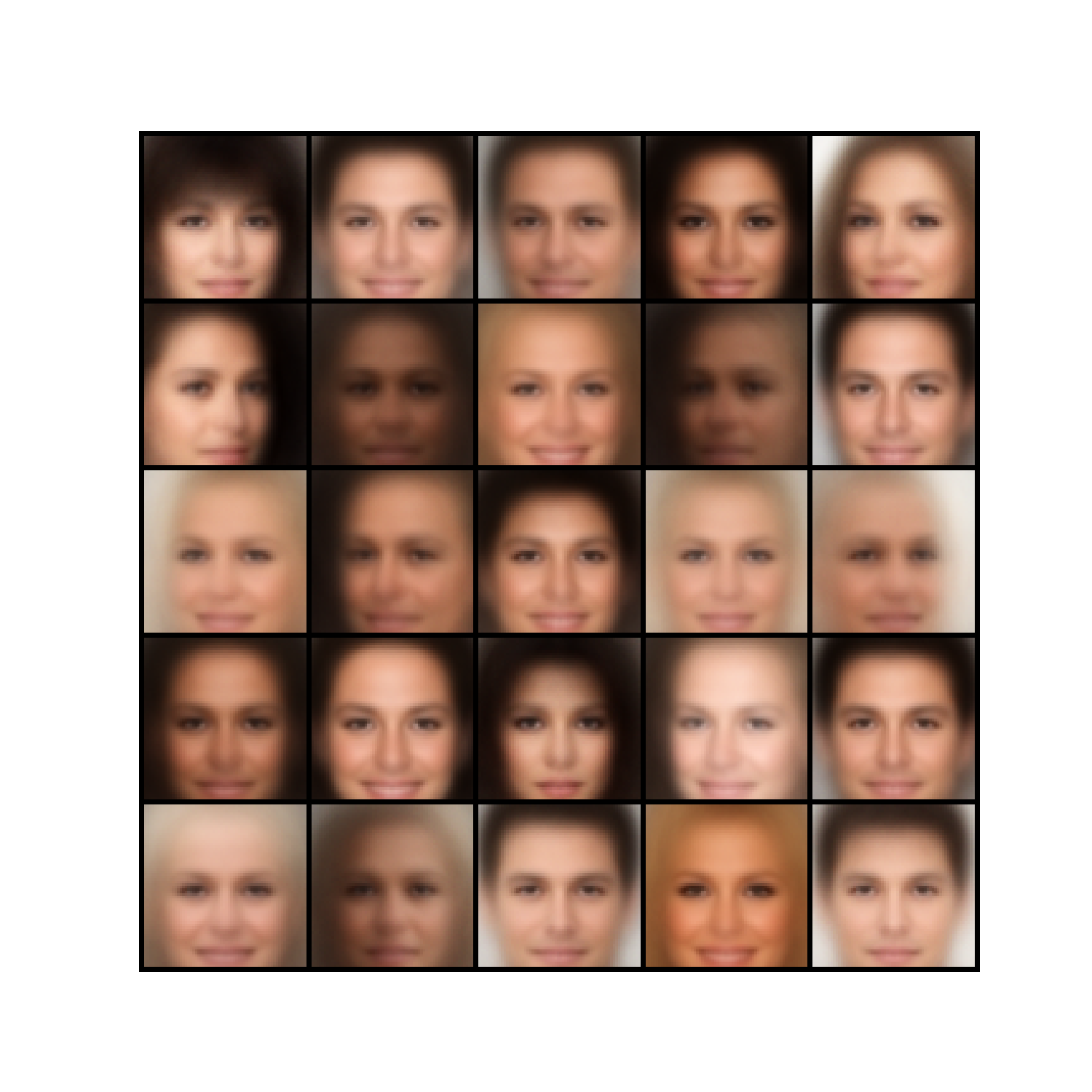}
         \caption{$\varepsilon = 0.01$}
     \end{subfigure}
     \hfill
     \begin{subfigure}{.5\textwidth}
         \centering
         \includegraphics[width=.5\textwidth]{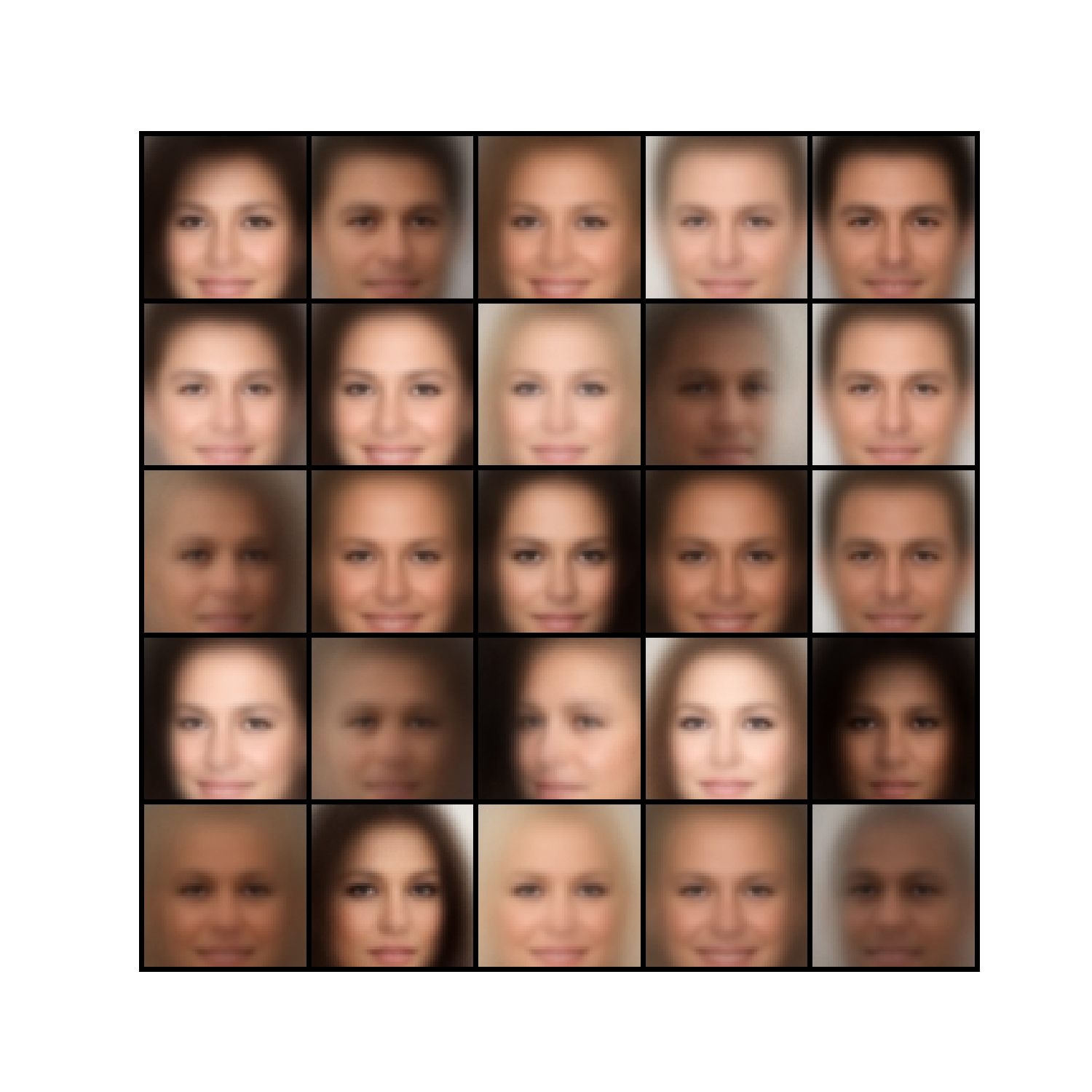}
         \caption{$\varepsilon = 1$}
     \end{subfigure}
     \hfill
     \begin{subfigure}{.5\textwidth}
         \centering
         \includegraphics[width=.5\textwidth]{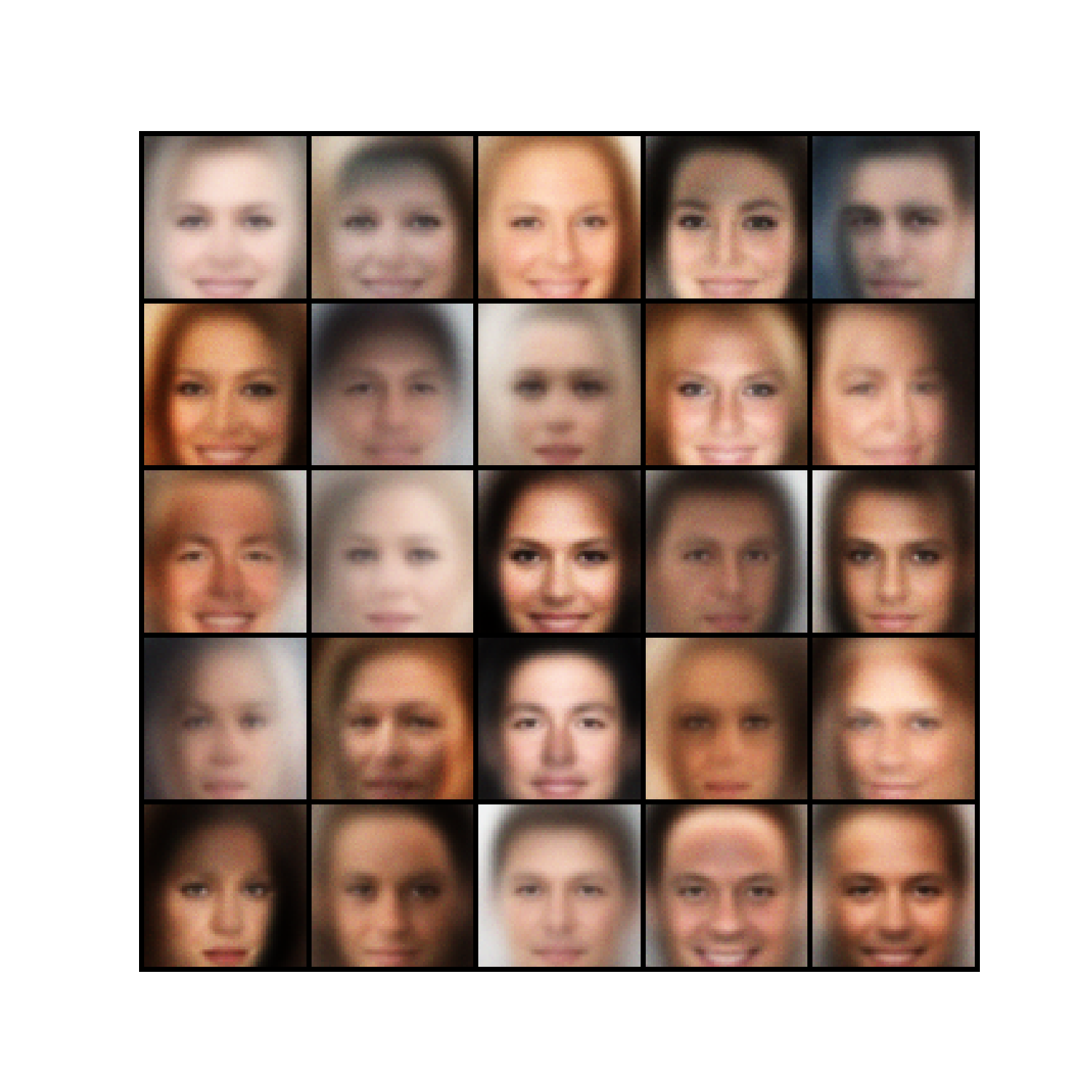}
         \caption{$\varepsilon = 10$}
     \end{subfigure}
     \hfill
     \begin{subfigure}{.5\textwidth}
         \centering
         \includegraphics[width=.5\textwidth]{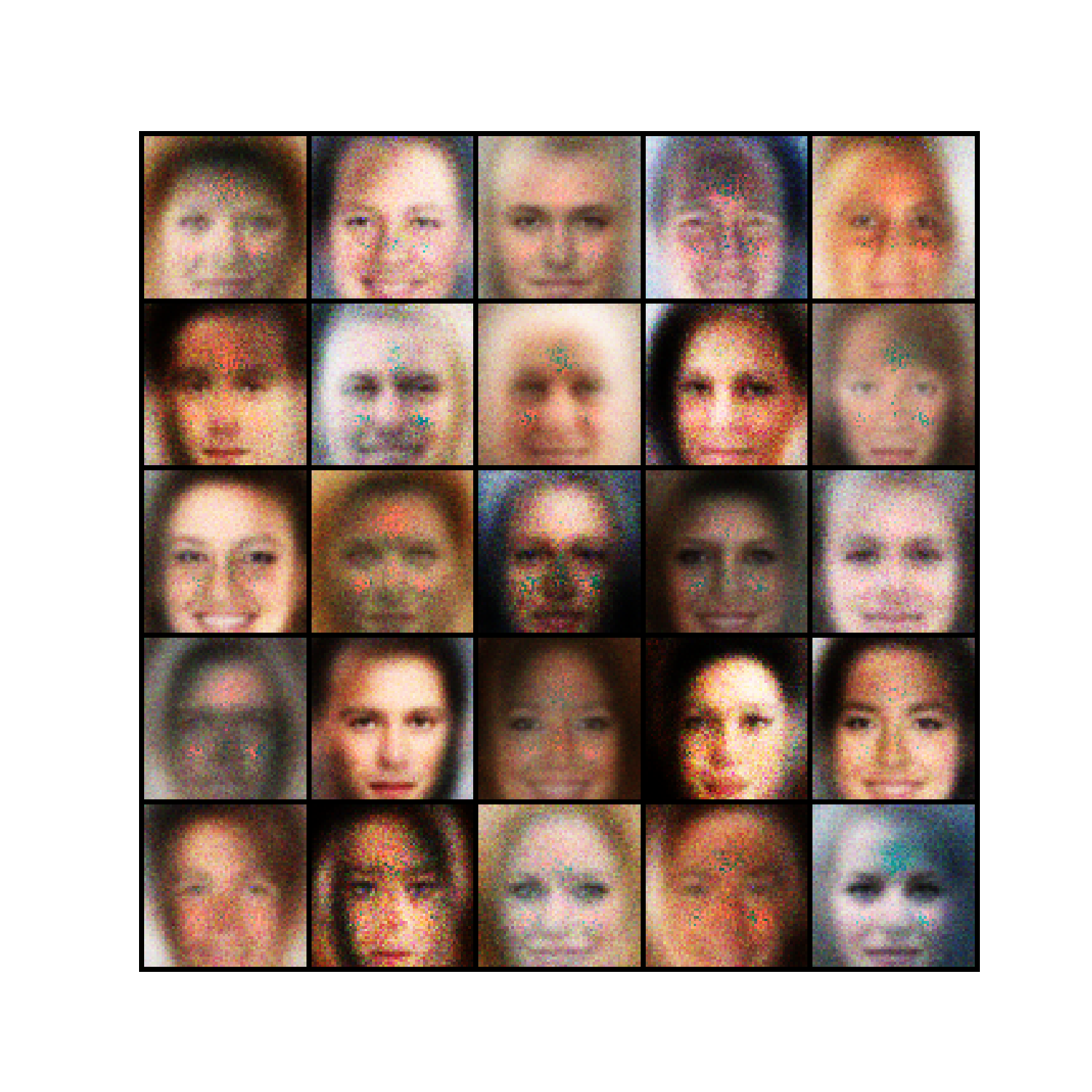}
         \caption{$\varepsilon = 100$}
     \end{subfigure}
        \caption{Images generated by a generator, which was trained using optimal discriminator dynamics, i.e.\ where the inner loop in Algorithm \ref{alg:WGAN-GP} is replaced by the true batch Sinkhorn distance, using the method described in \cite{learning_mb}. The problem of blurry, non-diverse images persists.}
        \label{fig:sinkhorn_batches}
\end{figure}

The authors in \citep{learning_mb} showed that the algorithm is successful in learning 2D euclidean data (8-mode Gaussian mixture), but they did not perform any experiments with image data.
Motivated by the failure of batch Wasserstein based losses and the connection to geometric $k$-medians described in the previous sections, we  trained their model on image data. The result of the  generator for CelebA is shown in Figure \ref{fig:sinkhorn_batches}. Like for the $c$-transform, the $(c,\varepsilon)$-transform and geometric $k$-medians clustering which all result in low batch Wasserstein distances, we obtain  non-diverse, blurry images. This demonstrates that mitigating discriminator suboptimality issues does not solve the problems with WGANs.

\subsection{Fundamental failures of the $L_2$-distance as a perceptual distance}
\label{sec: L2}
Why does a Wasserstein-like distance not produce  satisfactory images, even if issues related to sample complexity and  suboptimality of the discriminator are mitigated? Our hypothesis is that this is due to the fact that the Wasserstein distance (and the Sinkhorn divergence) between batches of images is based on the $L_2$-pixelwise distance between images. Similar conjecture has been made in \cite{chen2019gradual}. This can be seen from the primal formulation \eqref{eq:primal_formulation} of $W_1$ (in contrast to the dual Kantorovich-Rubenstein formulation \eqref{eq:dual_formulation}). 

For  finite point clouds $p,q$, we can represent them represented as a list of points and a list of probabilities for each point and allows to write \eqref{eq:primal_formulation} as 
\begin{align*}
    W_1(p,q) = \min_{\substack{\gamma \in \mathbb{R}^{ m \times n}}} \sum_{i,j} \gamma_{i,j} \norm{x_i - y_j}_2    \\
    \text{s.t. } \gamma \mathds{1} = P,\quad \gamma^T \mathds{1} = Q
\end{align*}
by \citep{ComputationalOT}.
Here, $\gamma_{i,j}$ denote the entries of the matrix $\gamma \in \mathbb{R}^{ m \times n}$, $x_i$ is the $i$-th image in $p$ and $y_j$ is $j$-th image in $q$. $P \in \mathbb{R}^n$ is such that $P_i $ is the probability mass at $x_i$ and $Q \in \mathbb{R}^m$ is such that $Q_i$ is the probability mass at $y_i$. 

When $p$ and $q$ both have supports of  size $n$, the Wasserstein distance satisfies
\begin{gather*}
    W_1(p,q) = \min_\sigma \sum_{i} \norm{x_i - y_{\sigma(i)}}_2,
\end{gather*}
where the minimisation is considered over all permutations $\sigma\colon \{1,\ldots,n\}\to \{1,\ldots,n\}$. 

When optimising the batch Wasserstein distance in the context of WGANs, we consider batches $p^\ast_n, p_n^\theta$ of size $n$, drawn from the target and generated probability densities $p^\ast, p^\theta$, respectively. Then, the Wasserstein distance between two batches $p^\ast_n$ and $p^\theta_ n$ is the  sum of pixel-wise $L_2$-distances, after the images in one of the batches are permuted.

\begin{figure}
    \centering
    \includegraphics[width=.48\textwidth]{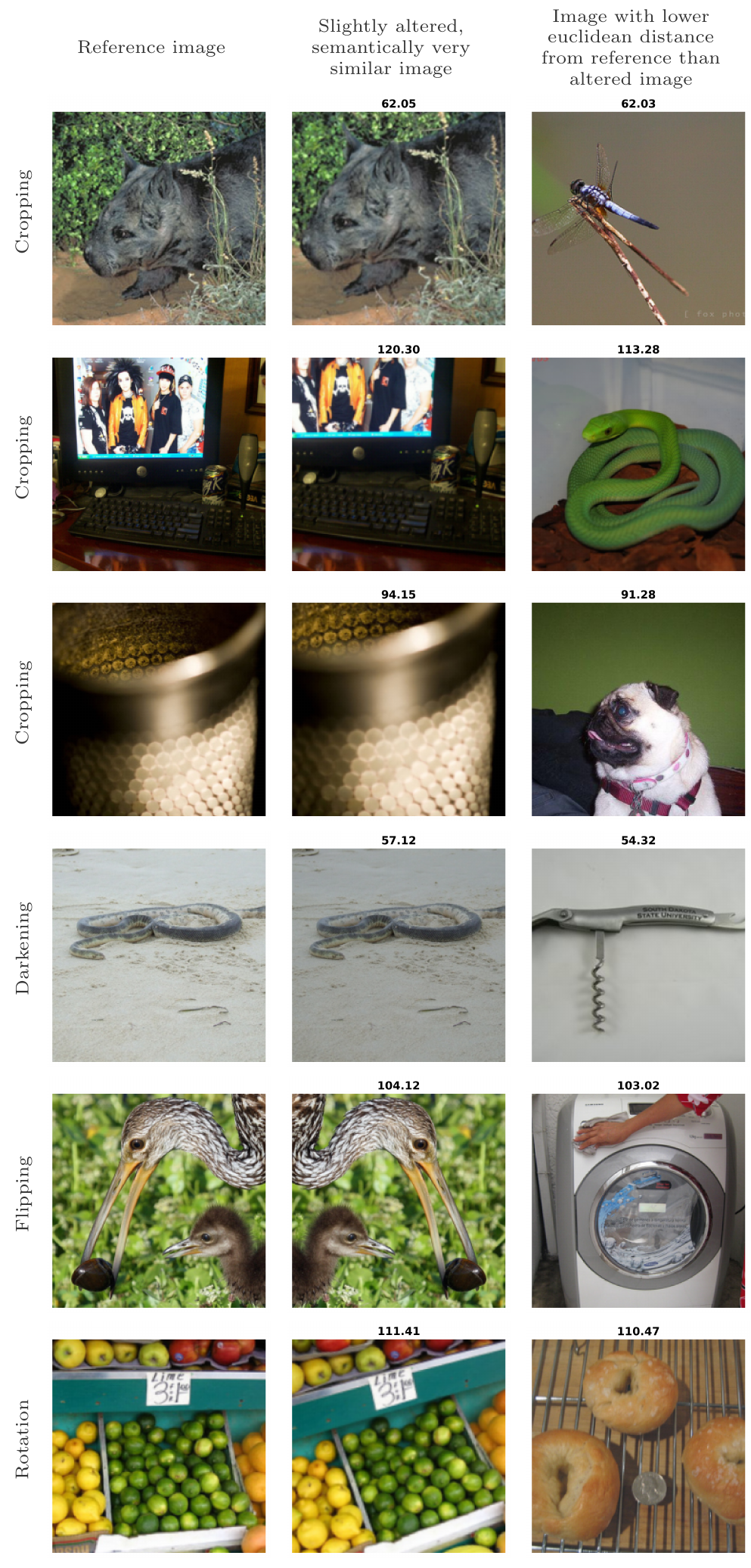}
    \caption{Examples where the $L_2$-distance is not a (semantically) meaningful distance between images. In the left column, a reference image from ImageNet \citep{imagenet} is shown. In the middle column, a slight alteration is applied, whose result a human observer would consider to be very close to the reference image. In the right column, another image from the ImageNet data set is displayed, which to the human observer is very different from the reference image, but which has a lower $L_2$-distance to the reference image than the altered image. The numbers above the images indicate the $L_2$-distance to the reference image.}
    \label{fig:L2bad}
\end{figure}

In terms of the $L_2$-distance, two images tend to be similar to one another, when the brightness values of their colour channels are similar. This does not necessarily meet our human perception of  two images being similar. For example, while the same person photographed under different lighting conditions would be assigned a low distance by a perceptual `human metric', pixelwise metrics would consider them dissimilar (which is also a reason for the specific centroids found in geometric $k$-medians clustering, cf.\ Figure \ref{fig:batches}). Wasserstein metrics on a space of distributions which are computed using pixelwise metrics on the respective sample space  thus always exhibit this phenomenon. Examples of failures of the euclidean metric to capture perceptual and semantic distances between images are shown in Figure \ref{fig:L2bad}. We gather absurd examples, which have a lower euclidean distance to some reference image than an image, which a human would consider only a slight variation of the reference image.
    
Moreover, since the Wasserstein distance and the Sinkhorn divergence are metrics based on pixelwise comparisons they disregard inductive bias encoded in convolutional neural networks, which captures spatial structure present in the image data. This inductive bias is essential for the success of deep learning models \cite{inductiveMontana}, \cite{cohen2017inductive}, \cite{deepImagePrior}. In fact, in our experiments  we  observed that methods based on accurate approximation of the batch Wasserstein distance perform similarly regardless of training with fully-connected or convolutional architectures. 

A possible explanation why WGAN-GP (and in fact other GANs) produce high-fidelity samples is precisely because they do not approximate any statistical divergence which is based on pixelwise metrics. Contrary to those, the training dynamics lead to a discriminator which captures the similarity and dissimilarity between samples better, since they are more flexible than the simple model of pixelwise distances.

\section{WGAN-GP in the space of GANs}
\label{sec: WGAN_space_of_GANs}
In the previous sections, we  established that the loss function of WGAN-GP does not approximate the Wasserstein distance in any meaningful sense. Moreover, accurate approximations of the Wasserstein distance are not even  desirable in minbatch-based training in the typical high-dimensional setup. This leads to a natural question: \textit{Why does WGAN-GP achieve such a good performance?} There are two possible answers in the literature which will be discussed in the following.

\subsection{Lipschitz regularisation}\label{sec:lipschitz_reg}
 It was suggested in \citep{DRAGAN} and \citep{Paths} that controlling the Lipschitz constant of the discriminator may improve GAN training regardless of the statistical distance used, and that the improved performance observed in WGAN-GP was simply due to the gradient penalty term and  not connected to the Wasserstein distance. 

\begin{figure}
  \centering
  \includegraphics[scale=.29]{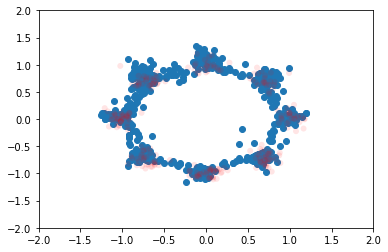}
  \includegraphics[scale=.29]{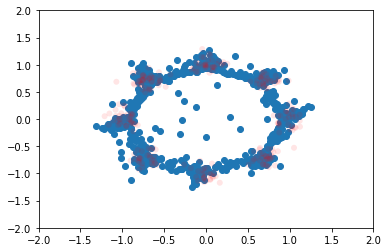}
  \caption{Two WGANs trying to learn an 8 mode Gaussian mixture. Red dots are a sampled from the target distribution and blue dots are sampled from the trained generator. Left:  WGAN with gradient penalty. Right:  WGAN with Lipschitz discriminator.}
  \label{lpzGAN1}
\end{figure}

\begin{figure}
  \centering
  \includegraphics[scale=.29]{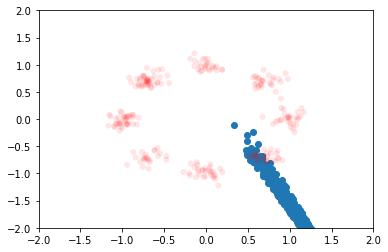}
  \includegraphics[scale=.29]{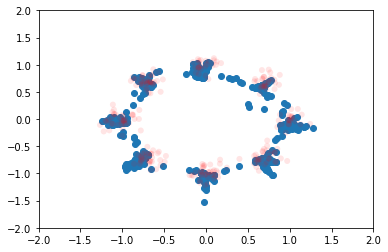}
  \caption{Left: Non saturating GAN with MLP discriminator. Right: Non saturating GAN with Lipschitz discriminator. Both trained using the same hyperparameters and computation time.}
  \label{lpzGAN2}
\end{figure}

Using the architectures described in \citep{LNets}, we trained a WGAN and a vanilla GAN \cite{GAN} with a multi-layer perceptron discriminator which is provably a 1-Lipschitz function. We compared this with vanilla GAN and WGAN-GP with an ordinary unconstrained multi-layer perceptron (MLP). In both cases, we observe an improved performance when using Lipschitz constrained discriminator, even though  1-Lipschitz functions have no theoretical connection to estimating the JS-divergence (which is implicitly used in the vanilla GAN). The results on a low-dimensional learning task are visualized in Figures \ref{lpzGAN1} and \ref{lpzGAN2}.

The importance of regularising the discriminator regardless of statistical distance used is further supported by \cite{schafer2020implicit}, where authors argue that if one does not impose regularity on the discriminator, it can achieve the maximal generator loss by exploiting visually imperceptible errors
in the generated images.

\subsection{Does the loss function even matter?}\label{sec:does_loss_fn_matter}
A large-scale study \citep{GANsEqual} in which different GAN loss functions were compared showed that GANs are very sensitive to their hyperparameter setup, and that no single loss function consistently outperforms the others. In particular it was shown that given the right hyperparameter configuration, vanilla GAN (refered to as NS GAN in this study) can achieve a comparable or better performance than WGAN-GP. The study examined many different flavors of GANs (MM GAN \citep{GAN}, NS GAN \citep{GAN}, DRAGAN \citep{DRAGAN}, WGAN \citep{WGAN}, WGAN GP \citep{WGAN-GP}, LS GAN \citep{mao2017squares}, BEGAN \citep{berthelot2017began}) on 4 benchmark data sets (MNIST, Fashion-MNIST, CIFAR10, CelebA). Each type of GAN had a different loss function and  was trained with 100 different hyperparameter configurations. The hyperparameter configurations were randomly selected from a wide range of possible values. The results of the study show that given the right hyperparameter configuration, vanilla GAN can achieve a performance very close to WGAN-GP. Therefore, the initially reported success of WGAN may be a result of a lucky hyperparmeter configuration and the precise loss function may be less relevant.

%\section{What about the regularisation?}
%To be done

\section{Conclusions}
In this article, we  examine several ways in which the theoretical foundations and the practical implementation of WGANs fundamentally differ. We argue that for good WGANs, the loss function typically does not approximate the Wasserstein distance well. Additionally, we gather evidence that good generators do not produce high-quality images \emph{despite} a poor approximation by the discriminator, but precisely \emph{because} it does not approximate the batch Wasserstein distance well.

In Section \ref{sec:Approximation of the optimal discriminator}, we show that the WGAN-GP loss function (which yields good generators) is not a meaningful approximation of $W_1$ due to the  poor approximation of the optimal discriminator. We point out that even if the optimal discriminator is approximated accurately via the $(c,\varepsilon)$-transform, the resulting loss function does not improve the performance of the model, as it exhibits problems such as 
`false' global minima. We show that these problems are significant in practice and point towards a possible reason, why this is the case (Section \ref{sec:finite_sample_approx}). Moreover, as we argue in Section \ref{sec:fundamental_problems}, even if we eliminate all  the \emph{practical} problems listed above, the Wasserstein distance  would still not be a suitable loss function, as the Wasserstein distance is based on a pixelwise distance in sample space which does not pose a perceptual distance metric. 

On the contrary, we assume that the flexibility of not  approximating a statistical metric based on  a pixelwise metric gives rise to better discriminator (critic) networks. We  partially attribute the claimed improvement in performance of WGAN over vanilla GAN to a control over the Lipschitz constant of the discriminator (Section \ref{sec:lipschitz_reg}). Moreover, some evidence suggests that the particular choice of the loss function does not  impose a significant impact on the model performance  (Section \ref{sec:does_loss_fn_matter}). Due to the above reasons, we conclude that the (true) Wasserstein distance is not a desirable loss function for GANs, and that WGANs should not be thought of as Wasserstein distance minimisers.

In future work, we would like to understand the role of the optimisation dynamics better. This concerns the two-player game between the generator and discriminator. Moreover, we would like to explore if WGANs can be modified so that they minimise a divergence based on a perceptually meaningful metric. A successful algorithm minimising optimal transport based distances needs to simultaneously address the following issues:
\begin{itemize}
    \item Sample complexity: Possibly by regularisation \cite{SinkhornComplexity} or sliced distances \cite{SlicedGAN}, \cite{Wsubspace}.
    \item Replacing the $L_2$-norm by a perceptually meaningful notion of distance: We intend to explore an optimal transport distance based on $L_2$ between VGG embeddings of the images in  future work.
    \item Approximation of optimal discriminator: Possibly by changing the optimisation algorithm.
\end{itemize}

\section*{Acknowledgements}
JS, LMK, CE and CBS thank Anton Mallasto for sharing the code used for his experiments.
JS, LMK and CBS acknowledge support from the Cantab Capital Institute for the Mathematics of Information. 
CE and CBS acknowledge support from the Wellcome Innovator Award RG98755.
LMK and CBS acknowledge support from the European Union Horizon 2020 research and innovation programmes under the Marie Sklodowska-Curie grant agreement No.\ 777826 (NoMADS).
JS additionally acknowledges the support from Aviva. 
LMK additionally acknowledges support from the Magdalene College, Cambridge (Nevile Research Fellowship). 
CBS additionally acknowledges support from the Philip Leverhulme Prize, the Royal Society Wolfson Fellowship, the EPSRC grants EP/S026045/1 and EP/T003553/1, EP/N014588/1, EP/T017961/1 and the Alan Turing Institute.

\bibliography{bibliography}
\bibliographystyle{icml2021}

\appendix
\section{Geometric k-medians}
\subsection{Formal definitions and proofs}
Following \cite{projection}, we define the nearest neighbour projection:
\begin{definition}[Nearest neighbour projection]
    Let $S \subseteq \mathbb{R}^d$ be a closed set. A nearest neighbour projection on $S$ is denoted by $\pi_S$ and defined as
    \begin{gather*}
        \pi_S(x) = \sum_{s \in S} s  1_{\substack{B_s(S)}}(x), \quad x\in \mathbb{R}^d,
    \end{gather*} 
    where $1_{\substack{B_s(S)}}$ is the indicator function of $B_s(S)$ and $B(S) :=(B_s(S))_{s\in S}$ is a Borel Voronoi partition of $\mathbb{R}^d$ such that 
    $B_s(S) \subseteq \{ x \in \mathbb{R}^d : \norm{x-s} = \min_{\substack{\tilde{s} \in S}} \norm{x - \tilde{s}} \}$. 
\end{definition}
Interested readers are referred to \cite{projection} for technical details.

\begin{remark}
    If $S$ is a finite set, then $\pi_S$  maps $x$ to a point in $S$ with minimal distance from $x$. If $S$ is not convex this point may be dependent on the choice of the Voronoi partition $B(S)$.
\end{remark}

\begin{definition} [Projection measure]
    Let $S \subseteq \mathbb{R}^d$ be a closed set and let $\pi_S$ be a nearest neighbour projection on $S$. Moreover, let $\rho$ be a probability distribution on $\mathbb{R}^d$. Then $\pi_S \rho$ denotes the distribution obtained as a push-forward of $\rho$ by $\pi_S$, i.e.\ $\pi_S \rho (A) = \rho(\pi_S^{-1}(A))$ for any $A\subset \mathbb{R}^d$.
\end{definition}

\begin{lemma} [\cite{canas}]
    \label{thm:exp_wass}
     Fix $ 1 \leq p < \infty$. Let $S \subseteq \mathbb{R}^d$ be a closed set,  and let $\rho$ be a probability distribution on $\mathbb{R}^d$ with finite p-th moment. We have
    \begin{gather*}
        \mathbb{E}_{x\sim \rho} d(x, S)^p = W_p(\rho, \pi_S \rho)^p,   
    \end{gather*}
    where $d(x,S)$ denotes the distance the smallest distance from $x$ to $S$ and $W_p$ is the Wasserstein-p metric.
\end{lemma}

\begin{lemma} [\cite{canas}]
    \label{thm: proj_min}
    Fix $ 1 \leq p < \infty$. Let $S \subseteq \mathbb{R}^d$ be a closed set,  and let $\rho$ be a probability distribution on $\mathbb{R}^d$ with finite p-th moment. For all probability distributions $\mu$ with finite p-th moment such that $\text{supp}(\mu) \subseteq S$, we have
    \begin{gather*}
        W_p(\rho, \mu) \geq W_p(\rho, \pi_S \rho).
    \end{gather*}
\end{lemma}

\begin{definition} [Geometric $k$ medians]
    Let $\mathcal X=\{x_i,~i=1,\ldots,n\}$ be a data set in $\mathbb{R}^d$. Given an arbitrary set $S=\{m_i, i=1,\ldots,k\}  \subseteq \mathbb{R}^d$, let  $S_i = \{ x \in \mathcal X : m_i = \argmin_{m \in S} \norm{m - x}_2 \}$, i.e.\ $S_i$ consists of the points in $\mathcal{X}$ for which the closest point in $S$ is $m_i$.  The \textit{geometric k-medians} is defined as the set $\hat{S}=\{\hat{m}_i,~i=1,\ldots,k\}$ such that
    \begin{align*}
        \hat{S} :&= \argmin_{S \subseteq \mathbb{R}^d : |S|=k} \sum_{i=1}^k \sum_{x \in \substack{S_i}} d(x,m_i) \\&= \argmin_{S \subseteq \mathbb{R}^d : |S|=k} \sum_{x \in \mathcal X} d(x,S).    
    \end{align*}
\end{definition}
\begin{theorem}\label{th:w1-kmedians}
    Let $\mathcal X=\{x_i,~i=1,\ldots,n\}$ be a data set in $\mathbb{R}^d$ and  let $\rho := \frac{1}{n} \sum_{i=1}^n \delta_{\substack{x_i}}$ be the associated empirical measure. Moreover, let $\hat{S}=\{\hat{m}_i,~i=1,\ldots,k\}$ be the geometric k-medians for $\mathcal X$ and let $\hat{S}_i := \{ x \in \mathcal X : \hat{m}_i = \argmin_{m \in \hat{S}} \norm{m - x}_2 \}$, i.e.\ the points in $\mathcal{X}$ for which the closest point in $\hat{S}$ is $\hat{m}_i$. Then,
    \begin{gather*}
        \pi_{\substack{\hat{S}}} \rho=\argmin_{\substack{\mu :\supp(\mu) = k }} W_1(\rho, \mu),
    \end{gather*}
    where $\pi_{\substack{\hat{S}}} = \frac{1}{n}\sum_{i=1}^n \delta_{\substack{\pi_{\substack{\hat{S}}}}(x_i)} = \sum_{i=1}^k \frac{|\hat{S}_i|}{n} \delta_{\substack{\hat{m}_i}}$.
\end{theorem}

Since $\pi_{\substack{\hat{S}}}\rho$ is an empirical measure concentrated on $\hat{S}$, Theorem \ref{th:w1-kmedians} states that the empirical distribution concentrated on geometric $k$-medians is the minimiser of the Wasserstein-1 distance between the empirical distribution of the data $\mathcal X$ and all distributions $\mu$ with  $\supp(\mu)=k$.

\begin{proof}
    We can proceed in close analogy with \cite{canas} where a connection between k-means and $W_2$ is examined.
    We have
    \begin{align*}
        \hat{S} &= \argmin_{S \subseteq \mathbb{R}^d : |S|=k} \sum_{x \in \mathcal X} d(x,S) 
        =  \argmin_{S \subseteq \mathbb{R}^d : |S|=k}\frac{1}{n} \sum_{x \in \mathcal X} d(x,S) \\
        &= \argmin_{S \subseteq \mathbb{R}^d : |S|=k} \mathbb{E}_{x\sim \rho} d(x, S) = \argmin_{S \subseteq \mathbb{R}^d \colon |S|=k} W_1(\rho, \pi_S \rho),
    \end{align*}
    where the last equality follows from Lemma \ref{thm:exp_wass}.
    Hence,
    \begin{gather*}
        \pi_{\substack{\hat{S}}}\rho = \argmin_{\substack{\pi_S \rho\colon S\subseteq \mathbb{R}^d, 
        |S|=k}} W_1(\rho, \pi_S \rho).
    \end{gather*}
    This means that $\pi_{\substack{\hat{S}}}\rho$ minimises the Wasserstein distance from $\rho$ among all measures which are projections on a set $S$ supported on $k$ points. On the other hand, we know that a projection measure always minimises the Wasserstein over all measures with support contained in a set $S$ (Lemma \ref{thm: proj_min}). Together, this implies that $\pi_{\substack{\hat{S}}}\rho$ minimises Wasserstein distance among all measures with support contained in a set $S$ supported on $k$ points. 
\end{proof}

\section{Differences to previous experiments}\label{sec:prevexp}
Here we explain crucial differences between three seemingly similar experiments, which explore different aspects of the approximation of the Wasserstein distance in WGANs. We compare our experiment described in Algorithm \ref{stanczuk} with an experiment of \citet{Pinetz} described in Algorithm \ref{pinetz} and an experiment of \citet{Mallasto} described in Algorithm \ref{mallasto}.

\begin{algorithm2e}
    \SetAlgoLined
    \DontPrintSemicolon
    \For{$N$ iterations}{
        Sample a batch $p_n^\ast$ from $p^\ast$ \;
        Sample a batch $p_n^\theta$ from $p^\theta$ \;
        Ascent step on $D_\alpha$ wrt. $\mathcal V(D_\alpha, p_n^\ast, p_n^\theta) - \lambda \mathcal{R}(D_\alpha, p^\ast_n, p^\theta_n)$\;
    }
    $W_1^D(p^\ast,p^\theta) \gets \mathbb{E}_{x \sim p^\ast}[D_\alpha(x)] - \mathbb{E}_{x \sim p^\theta}[D_\alpha(x)]$ \;
    $W_1(p^\ast,p^\theta) \gets $ Solution of LP for $p^\ast,p^\theta$\;
    Compare $W_1^D(p^\ast,p^\theta)$ and $W_1(p^\ast,p^\theta)$
\caption{Approximation Stanczuk et al}
\label{stanczuk}
\end{algorithm2e}

\begin{algorithm2e}
    \SetAlgoLined
    \DontPrintSemicolon
    \For{$N$ iterations}{
        Ascent step on $D_\alpha$ wrt. $\mathcal V(D_\alpha, p^\ast,p^\theta)$\;
    }
    $W_1^D(p^\ast,p^\theta) \gets \mathbb{E}_{x \sim p^\ast}[D_\alpha(x)] - \mathbb{E}_{x \sim p^\theta}[D_\alpha(x)]$ \;
    $W_1(p^\ast,p^\theta) \gets $ Solution for LP for $p^\ast,p^\theta$\;
    Compare $W_1^D(p^\ast,p^\theta)$ and $W_1(p^\ast,p^\theta)$
\caption{Approximation Pinetz et al}
\label{pinetz}
\end{algorithm2e}

\begin{algorithm2e}
    \SetAlgoLined
    \DontPrintSemicolon
    \For{$N$ iterations}{
        Sample a batch $p_n^\ast$ from $p^\ast$ \;
        Sample a batch $p^\theta_n$ from $p^\theta$ \;
        Ascent step on $D_\alpha$ wrt. $\mathcal V(D_\alpha, p_n^\ast, p^\theta_n)$\;
    }
    \For {$M$ iterations}{
        Sample a batch $p_n^\ast$ from $p^\ast$ \;
        Sample a batch $p^\theta_n$ from $p^\theta$ \;
        $W_1^D(p^\ast,p^\theta) \gets \mathbb{E}_{x \sim p^\ast}[D_\alpha(x)] - \mathbb{E}_{x \sim p^\theta}[D_\alpha(x)]$ \;
        $W_1(p_n^\ast,p^\theta_n) \gets $ Solution for LP for $p^\ast,p^\theta$\;
        Compare $W_1^D(p_n^\ast,p^\theta_n)$ and $W_1(p_n^\ast,p^\theta_n)$
    } 
\caption{Approximation Mallasto et al}
\label{mallasto}
\end{algorithm2e}

\citet{Pinetz} take two small batches ($n=500$) of CIFAR-10 images and train the discriminator using to WGAN-GP loss function as described in the Algorithm \ref{pinetz}. Then the authors check how well $\mathcal V(D_\alpha, p^\ast,p^\theta)$ approximates $W_1(p^\ast,p^\theta)$, after the discriminator has been trained.  We emphasise that the training is done using full batches and not mini-batches, and that the measures $p^\ast,p^\theta$ consider small sample sizes ($500$ samples). Contrary to our experiment it does not examine the mini-batch dynamics used in WGAN-GP, where algorithm tries to approximate the distributional Wasserstein distance $W_1(p^\ast,p^\theta)$ based on many small batches $p^\ast_n$ and $p_n^\theta$. 

In contrast we pick two large finitely supported distributions $p^\ast$ and $p^\theta$, each consists of $10$K images  from CIFAR-10 \cite{CIFAR}.  Then we sample mini-batches of size $n=64$ from $p^\ast,p^\theta$ and maximise $\loss_D(\alpha)$. This is exactly the same procedure as in the WGAN-GP training in Algorithm~\ref{alg:WGAN-GP} except that  both measures are static (as if the generator in Algorithm \ref{alg:WGAN-GP} was frozen).
 
The experiment of \cite{Mallasto}  checks the quality of minbatch estimator rather than the oracle estimator, i.e.\ how well on average $\mathcal V(D_\alpha, p_n^\ast, p^\theta_n)$ approximates $W_1(p_n^\ast,p^\theta_n)$ (after the discriminator has been trained) and not how well $\mathcal V(D_\alpha, p_n^\ast, p^\theta_n)$ approximates $W_1(p^\ast,p^\theta)$ from which the batches are drawn.

\end{document}